\documentclass[letterpaper]{article} %
\usepackage{colorai25hyperref}  %
\usepackage{times}  %
\usepackage{helvet}  %
\usepackage{courier}  %
\usepackage[hyphens]{url}  %
\usepackage{graphicx} %
\usepackage{natbib}  %
\usepackage{caption} %
\usepackage{palette}
\usepackage{cirtikz}

\newcommand{\emoji}[1]{%
  \begingroup\normalfont\Large\raisebox{-1pt}{\includegraphics[height=\fontcharht\font`\B]{#1-emoji.png}}\endgroup
}

\usepackage{newfloat}
\usepackage{listings}

\usepackage{multicol}
\usepackage{hyperref}
\usepackage{aaai25researchpack}
\hypersetup{
colorlinks=true,linkcolor=lacamoil5!75!black,citecolor=lacamdarklilac4,urlcolor=pinkfluo!70!black
}

\renewcommand{\paragraph}[1]{\vspace{1pt}\noindent\textbf{#1}}

\DeclareCaptionStyle{ruled}{labelfont=normalfont,labelsep=colon,strut=off} %
\floatstyle{ruled}
\newfloat{listing}{tb}{lst}{}
\floatname{listing}{Listing}
\title{On Faster Marginalization with Squared Circuits via Orthonormalization}

\author {
    Lorenzo Loconte\textsuperscript{\emoji{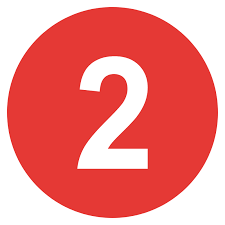}}\qquad
    Antonio Vergari\textsuperscript{\emoji{circled-two}}
}
\affiliations {
    \textsuperscript{\emoji{circled-two}} School of Informatics, University of Edinburgh, UK\\[3pt]
    l.loconte@sms.ed.ac.uk,
    avergari@ed.ac.uk
}

\begin{document}

\maketitle

\begin{abstract}

Squared tensor networks (TNs) and their generalization as parameterized computational graphs -- \emph{squared circuits} -- have been recently used as expressive distribution estimators in high dimensions.
However, the squaring operation introduces additional complexity when marginalizing variables or computing the partition function,
which hinders their usage in machine learning applications.
Canonical forms of popular TNs are parameterized via unitary matrices as to simplify the computation of particular marginals, but cannot be mapped to general circuits since these might not correspond to a known TN.
Inspired by TN canonical forms, we show how to parameterize squared circuits to ensure they encode already normalized distributions.
We then use this parameterization to devise an algorithm to compute any marginal of squared circuits that is more efficient than a previously known one.
We conclude by formally showing the proposed parameterization comes with no expressiveness loss for many circuit classes.

\end{abstract}

\section{Introduction}\label{sec:introduction}

Tensor networks (TNs) are low-rank tensor factorizations often used to compactly represent high-dimensional probability distributions, both in quantum physics \citep{orus2013practical,biamonte2017tensor} and in ML \citep{stoudenmire2016supervised,glasser2018from,cheng2019tree,glasser2019expressive,novikov2021ttde}.
A TN factorizing a complex function $\psi$ over a set of variables $\vX = \{X_i\}_{i=1}^d$ having domain $\dom(\vX)$ can be used to represent a probability distribution via modulus squaring, i.e., $p(\vX) = Z^{-1} |\psi(\vX)|^2 = Z^{-1} \psi(\vX)\psi(\vX)^*$, where $(\ \cdot\ )^*$ denotes the complex conjugation, and $Z = \int_{\dom(\vX)} |\psi(\vx)|^2 \mathrm{d}\vx$ is the partition function.

Recently, \citet{loconte2023turn,loconte2024relationship} showed that TNs can be generalized into deep computational graphs called \emph{circuits} \citep{choi2020pc}.
This is done by casting tensor contraction operations into layers of sums and products whose feed-forward evaluation corresponds to  a complete contraction to evaluate $\psi$.
The language of circuits offers the opportunity to flexibly build novel TN structures by just stacking layers of sums and products as ``Lego blocks'' \citep{loconte2024relationship}; include different basis input functions, and offering a seamless integration with deep learning \citep{shao2022conditional,gala2024pic,gala2024scaling}.
Furthermore, casting TNs as circuits provides conditions as to compose them and compute many probabilistic reasoning tasks in closed-form, such as expectations and information-theoretic measures \citep{vergari2021compositional}, which is crucial, e.g., for reliable neuro-symbolic AI \citep{ahmed2022semantic,zhang2023tractable}.
This is done with \emph{probabilistic circuits} (PCs), circuits encoding probability distributions, that are classically restricted to have positive parameters only --
also called \emph{monotonic} PCs \citep{shpilka2010open}.

One can increase the expressiveness of PCs by equipping them with complex parameters, \textit{squaring} them \citep{loconte2024subtractive}, similarly to TNs, or even more by mixing squared PCs \citep{loconte2024sos}.
Differently from classical monotonic PCs, which are not squared, the squared PCs require  additional computation to be normalized, i.e., to compute $Z$, which is quadratic in the circuit size.
This computational overhead carries over to computing marginals and hinders their application in a number of tasks such as lossless compression \citep{liu2022lossless} and sampling \citep{loconte2024subtractive}, where performing fast marginalization is crucial.

In this paper,
we show that the solution to
this inefficiency of squared PCs
comes from the literature of TNs, where \emph{canonical forms} are  adopted in order to simplify the computation of
probabilities \citep{schollwoeck2010density}.
For instance, instead of computing $Z$ explicitly in the case of a matrix-product state (MPS) TN \citep{perez2006mps}, a canonical form ensures $|\psi(\vX)|^2$ is an already-normalized probability distribution, i.e., $Z = 1$.
In practice, canonical forms are obtained by parameterizing a TN using unitary matrices, i.e., matrices $\vA\in\bbC^{n\times n}$ satisfying $\vA^\dagger\vA = \vA\vA^\dagger = \vI_n$, where $\vI_n$ denotes the identity matrix of size $n$.
{For rectangular matrices, \emph{semi-unitary} matrices $\vA\in\bbC^{m\times n}$ are considered, i.e., matrices satisfying $\vA^\dagger\vA = \vI_n$ if $m > n$ or $\vA\vA^\dagger = \vI_m$ if $m < n$.}
Under this view, computing $Z$ simplifies into operations over identity matrices.
However, computing different marginals over different TN structures requires a different canonical form and these cannot be immediately translated to squared PCs, because the  latter allows us to build factorizations that might not correspond to any known TN.
This begs the question: \emph{how can we parameterize squared circuits as to be already normalized and allow us to accelerate the computation of} any \emph{marginal}? We answer it in the following.

\paragraph{Contributions.}
We derive sufficient conditions via orthonormal parameterizations to ensure that squared PCs are already normalized (\cref{sec:normalized-squared}).
These conditions are based on semi-unitary matrices, similarly to canonical forms in TNs, but defined within the language of tensorized circuits instead.
Then, by leveraging squared orthonormal PCs, we present a general algorithm to compute any marginal that can be much more efficient than the previously known algorithm for squared PCs which required a quadratic increase in complexity instead (\cref{sec:tighter-marginalization}).
Our algorithm, which exploits the concept of variable dependencies of the circuit layers, can be used to speed up the computation or arbitrary marginals for TNs as well.
Finally, we show how the proposed parameterization can be enforced efficiently in squared PCs, thus theoretically guaranteeing no loss of expressiveness for certain circuit families (\cref{sec:question-expressiveness}).

\section{Circuits and Squared Circuits}\label{sec:squared-circuits}

We start by defining circuits in a tensorized formalism \citep{vergari2019visualizing,loconte2024relationship}.
\begin{defn}[Tensorized circuit]
    \label{defn:tensorized-circuit}
    A \emph{tensorized circuit} $c$ is a parameterized computational graph encoding a function $c(\vX)$ and comprising of three kinds of layers: \emph{input}, \emph{product} and \emph{sum}.
    A layer $\vell$ is a vector-valued function defined over variables $\uscope(\vell)$, called \textit{scope}, and every non-input layer receives the outputs of other layers as input, denoted as $\inscope(\vell)$.
    The scope of each non-input layer is the union of the scope of its inputs.
    The three kinds of layers are defined as follows:
    \begin{itemize}
        \item Each \emph{input layer} $\vell$ has scope $X\in\vX$ and computes a collection of $K$ input functions $\{f_i\colon\dom(X)\to\bbC\}_{i=1}^K$, i.e., $\vell$ outputs a $K$-dimensional vector.
        \item Each \emph{product layer} $\vell$ computes either an element-wise (or Hadamard) or Kronecker product of its $N$ inputs, i.e., $\odot_{i=1}^N \vell_i(\uscope(\vell_i))$ or $\otimes_{i=1}^N \vell_i(\uscope(\vell_i))$, respectively.
        \item A sum layer $\vell$ with $\vell_1$ as input, is parameterized by $\vW\in\bbC^{K_1\times K_2}$ and computes the matrix-vector product $\vell(\uscope(\vell)) = \vW \vell_1(\uscope(\vell_1))$.
    \end{itemize} 
\end{defn}

Each non-input layer is a vector-valued function, but each entry it computes is a scalar-valued function computed over certain entries of its
input vectors.
We denote as $\lsize(\vell)$ the number of scalar inputs to each scalar function encoded in $\vell$.
That is, an Hadamard layer consists of $K$ scalar functions each computing the product of $N$ other scalars, thus $\lsize(\vell) = NK$.
A Kronecker layer consists of $K^N$ scalar functions each computing the product of $N$ other scalars, i.e., $\lsize(\vell) = NK^N$.
Finally, a sum layer consists of $K_1$ scalar functions each receiving input from $K_2$ other scalars and computing a linear combination, i.e., $\lsize(\vell) = K_1K_2$.
The size of a layer is also its computational complexity.

A tensorized PC is a tensorized circuit $c$ computing non-negative values, i.e., for any $\vx$ we have $c(\vx)\geq 0$.
Thus, a PC $c$ encodes a probability distribution $p(\vx) = Z^{-1} c(\vx)$.
A PC $c$ supports tractable marginalization of any variable subset \citep{choi2020pc} if (i) the functions encoded by the input layers can be integrated efficiently and (ii) it is \emph{smooth} and \emph{decomposable}.

\begin{defn}[Layer-wise smoothness and decomposability \citep{darwiche2002knowledge,loconte2024relationship}]
    \label{defn:smoothness-decomposability}
    A tensorized circuit over variables $\vX$ is \emph{smooth} if for every sum layer $\vell$, its inputs depend on all the same variables, i.e., $\forall \vell_i,\vell_j\in\inscope(\vell)\colon \uscope(\vell_i) = \uscope(\vell_j)$.
    It is \emph{decomposable} if for every product layer $\vell$ in it, its inputs depend on disjoint scopes, i.e., $\forall \vell_i,\vell_j\in\inscope(\vell),i\neq j\colon \uscope(\vell_i)\cap\uscope(\vell_j)=\emptyset$.
\end{defn}

Since sum layers can have only one input in \cref{defn:tensorized-circuit}, circuits are ensured smooth, but not necessarily decomposable.
Popular TNs like MPS \citep{perez2006mps} and tree TNs \citep{shi2006tree,cheng2019tree} are special cases of smooth and decomposable circuits having a particular tree structure, and use Hadamard and Kronecker layers, respectively \citep{loconte2024relationship}.
However, \cref{defn:tensorized-circuit} allows us to build different factorizations by connecting layers, e.g., mix both Hadamard and Kronecker layers, include different input functions, and share parameters.

\citet{loconte2024sos} showed one can learn PCs with complex parameters by \emph{squaring} circuits.
Given a circuit $c$ that outputs complex numbers, a squared PC $c^2$ is obtained by multiplying $c$ and its complex conjugate $c^*$, i.e., $c^2(\vx) = |c(\vx)|^2 = c(\vx) c^*(\vx)$.
Thus, a squared PC encodes a distribution $p(\vx) = Z^{-1} |c(\vx)|^2$, where $Z = \int_{\dom(\vX)} |c(\vx)|^2 \mathrm{d}\vx$.
Computing $Z$ tractably requires representing $c^2$ as another decomposable circuit, which can be done if $c$ is \emph{structured-decomposable} \citep{vergari2021compositional}, i.e., each product layer factorizes its scope to its inputs, and the collection of such factorizations forms a tree \citep{pipatsrisawat2008new}.
As detailed in \cref{app:squaring-circuits}, $c^2$ can be built from $c$ by retaining its structure and by quadratically increasing the size of layers.
Thus, computing $Z$ requires time $\calO(LS^2)$, where $L$ is the number of layers and $S$ is the maximum layer size in $c$, i.e., $S = \max_{\vell\in c}\{\lsize(\vell)\}$.
In general, marginalizing any variable subset still requires time $\calO(LS^2)$ \citep{loconte2024subtractive}.

Instead, monotonic PCs can be parameterized by (i) using distributions as input functions (e.g., Gaussian), and
(ii) normalizing each parameter matrix row, i.e., for each sum layer parameterized by $\vW\in\bbR_+^{K_1\times K_2}$ we have that $\forall i\in [K_1]$ $\sum_{j=1}^K w_{ij} = 1$, where we denote $[n]=\{1,\ldots,n\}$.
The key advantage of (i,ii) is that that the resulting PC encodes an already-normalized distribution.
Moreover, marginalizing any variable subset requires time $\calO(LS)$, as they are not squared.
Note that (ii) is not restrictive, as it can be enforced efficiently using an algorithm by \citet{peharz2015theoretical}.

For squared PCs, \emph{what are the sufficient conditions ensuring they are already normalized?}
We present them next, and in \cref{sec:tighter-marginalization} we show how they lead to an algorithm to compute any marginal that can be much more efficient.

\begin{figure*}[!t]
\begin{subfigure}{0.31\linewidth}
\scalebox{0.67}{
\begin{tikzpicture}[cirtikz, minimum height=45pt]
    \node (i1) [fill=yellow3, anchor=north west] {\Gaussian};
    \node (i2) [fill=tomato3, anchor=north, below=15pt of i1.south] {\Gaussian};
    \node (i3) [fill=deliveroo, anchor=north, below=15pt of i2.south] {\Gaussian};
    \node (i4) [fill=lacamlilac, anchor=north, below=15pt of i3.south] {\Gaussian};
    \node (p1) [fill=lacamgold4, anchor=west, right=20pt of i1.east] at ($(i1.east)!0.5!(i2.east)$) {\Odot};
    \node (p2) [fill=petroil4, anchor=west, right=20pt of i3.east] at ($(i3.east)!0.5!(i4.east)$) {\Odot};
    \node (r1) [fill=lacamgold4, anchor=west, right=20pt of p1.east] {\Sum};
    \node (r2) [fill=petroil4, anchor=west, right=20pt of p2.east] {\Sum};
    \node (q) [fill=olive4, anchor=west, right=20pt of r1.east] at ($(r1.east)!0.5!(r2.east)$) {\Odot};
    \node (s) [fill=olive4, minimum width=20pt, minimum height=20pt, anchor=west, right=20pt of q.east] {\Sum};
\begin{pgfonlayer}{background}
    \node (non-marg-bg) [ultra thick, draw=tomato3!40!bgrey3, dotted, anchor=north west, minimum width=108pt, minimum height=112pt, outer sep=-5pt] {};
    \node (marg-bg) [ultra thick, draw=deliveroo!40!bgrey3, dotted, below=15pt of non-marg-bg.south west, anchor=north west, minimum width=108pt, minimum height=112pt, outer sep=-5pt] {};
    \node (mixed-bg) [ultra thick, draw=olive4!60!bgrey3, dotted, right=16pt of non-marg-bg.east, anchor=west, minimum width=70pt, minimum height=54pt, outer sep=-5pt] at ($(non-marg-bg.east)!0.5!(marg-bg.east)$) {};
    \draw [-, bcedge] (i1.south east) -- (i2.north east);
    \draw [-, bcedge] (i1.north east) -- (p1.north west);
    \draw [-, bcedge] (i2.south east) -- (p1.south west);
    \draw [-, bcedge] (p1.north east) -- (r1.north west);
    \draw [-, bcedge] (p1.south east) -- (r1.south west);
    \draw [-, bcedge] (i3.south east) -- (i4.north east);
    \draw [-, bcedge] (i3.north east) -- (p2.north west);
    \draw [-, bcedge] (i4.south east) -- (p2.south west);
    \draw [-, bcedge] (p2.north east) -- (r2.north west);
    \draw [-, bcedge] (p2.south east) -- (r2.south west);
    \draw [-, bcedge] (q.north east) -- (s.north west);
    \draw [-, bcedge] (q.south east) -- (s.south west);
    \draw [-, bcedge] (r1.south east) -- (r2.north east);
    \draw [-, bcedge] (r1.north east) -- (q.north west);
    \draw [-, bcedge] (r2.south east) -- (q.south west);
    \draw [o-, bcedge, fill=bdark, shorten <=-4pt, shorten >=-15pt] ($(p1.east)!0.5!(r1.west)$) -- ($(p1.north east)!0.5!(r1.north west)$) node [yshift=22pt] {\large\color{black} $\vW_2$};
    \draw [o-, bcedge, fill=bdark, shorten <=-4pt, shorten >=-15pt] ($(p2.east)!0.5!(r2.west)$) -- ($(p2.north east)!0.5!(r2.north west)$) node [yshift=22pt] {\large\color{black} $\vW_3$};
    \draw [o-, bcedge, fill=bdark, shorten <=-4pt, shorten >=-15pt] ($(q.east)!0.5!(s.west)$) -- ($(q.north east)!0.5!(s.north west)$) node [yshift=22pt] {\large\color{black} $\vW_1$};
\end{pgfonlayer}
\begin{pgfonlayer}{foreground}
    \node (i1l) [left=8pt of i1.west] {\large $f_1(X_1)$};
    \node (i2l) [left=8pt of i2.west] {\large $f_2(X_2)$};
    \node (i3l) [left=8pt of i3.west] {\large $f_3(X_3)$};
    \node (i4l) [left=8pt of i4.west] {\large $f_4(X_4)$};
    \node (phi-evidence) [text height=6pt, text width=10pt, anchor=north west, right=8pt of r1.north east] {\LARGE $\phi_{\vY}$};
    \node (phi-marginalized) [text height=6pt, text width=10pt, anchor=south west, right=8pt of r2.south east] {\LARGE $\phi_{\vZ}$};
    \node (phi-mixed) [text height=6pt, text width=10pt, anchor=north west, below=-4pt of mixed-bg.south] {\LARGE $\phi_{\vY,\vZ}$};
\end{pgfonlayer}
\end{tikzpicture}
\vspace{-1em}
}
\end{subfigure}%
\begin{subfigure}{0.42\linewidth}
\scalebox{0.67}{
\begin{tikzpicture}[cirtikz, minimum width=45pt, minimum height=45pt]
    \node (i1) [fill=yellow3, anchor=north west] {\Gaussian};
    \node (i2) [fill=tomato3, anchor=north, below=15pt of i1.south] {\Gaussian};
    \node (i3) [fill=deliveroo, anchor=north, below=15pt of i2.south] {\Gaussian};
    \node (i4) [fill=lacamlilac, anchor=north, below=15pt of i3.south] {\Gaussian};
    \node (p1) [fill=lacamgold4, anchor=west, right=20pt of i1.east] at ($(i1.east)!0.5!(i2.east)$) {\Odot};
    \node (p2) [fill=petroil4, anchor=west, right=20pt of i3.east] at ($(i3.east)!0.5!(i4.east)$) {\Odot};
    \node (r1) [fill=lacamgold4, anchor=west, right=20pt of p1.east] {\Sum};
    \node (r2) [fill=petroil4, anchor=west, right=20pt of p2.east] {\Sum};
    \node (q) [fill=olive4, anchor=west, right=20pt of r1.east] at ($(r1.east)!0.5!(r2.east)$) {\Odot};
    \node (s) [fill=olive4, minimum width=20pt, minimum width=20pt, minimum height=20pt, anchor=west, right=20pt of q.east] {\Sum};
\begin{pgfonlayer}{foreground}
    \node (i1l) [left=0pt of i1.west] {\large \shortstack{$f_1(X_1)$ \\ $\otimes$ \\ $f_1(X_1)^*$}};
    \node (i2l) [left=0pt of i2.west] {\large \shortstack{$f_2(X_2)$ \\ $\otimes$ \\ $f_2(X_2)^*$}};
    \node (i3l) [left=3pt of i3.west] {\large \shortstack{$\int f_3(x_3)$ \\ $\otimes$ \\ $f_3(x_3)^* \mathrm{d}x_3$}};
    \node (i4l) [left=3pt of i4.west] {\large \shortstack{$\int f_4(x_4)$ \\ $\otimes$ \\ $f_4(x_4)^* \mathrm{d}x_4$}};
\end{pgfonlayer}
\begin{pgfonlayer}{background}
    \draw [-, bcedge] (i1.south east) -- (i2.north east);
    \draw [-, bcedge] (i1.north east) -- (p1.north west);
    \draw [-, bcedge] (i2.south east) -- (p1.south west);
    \draw [-, bcedge] (p1.north east) -- (r1.north west);
    \draw [-, bcedge] (p1.south east) -- (r1.south west);
    \draw [-, bcedge] (i3.south east) -- (i4.north east);
    \draw [-, bcedge] (i3.north east) -- (p2.north west);
    \draw [-, bcedge] (i4.south east) -- (p2.south west);
    \draw [-, bcedge] (p2.north east) -- (r2.north west);
    \draw [-, bcedge] (p2.south east) -- (r2.south west);
    \draw [-, bcedge] (q.north east) -- (s.north west);
    \draw [-, bcedge] (q.south east) -- (s.south west);
    \draw [-, bcedge] (r1.south east) -- (r2.north east);
    \draw [-, bcedge] (r1.north east) -- (q.north west);
    \draw [-, bcedge] (r2.south east) -- (q.south west);
    \draw [o-, bcedge, fill=bdark, shorten <=-4pt, shorten >=-15pt] ($(p1.east)!0.5!(r1.west)$) -- ($(p1.north east)!0.5!(r1.north west)$) node [yshift=22pt] {\large\color{black} $\vW_2\otimes\vW_2^*$};
    \draw [o-, bcedge, fill=bdark, shorten <=-4pt, shorten >=-15pt] ($(p2.east)!0.5!(r2.west)$) -- ($(p2.north east)!0.5!(r2.north west)$) node [yshift=22pt] {\large\color{black} $\vW_3\otimes\vW_3^*$};
    \draw [o-, bcedge, fill=bdark, shorten <=-4pt, shorten >=-15pt] ($(q.east)!0.5!(s.west)$) -- ($(q.north east)!0.5!(s.north west)$) node [yshift=22pt] {\large\color{black} $\vW_1\otimes\vW_1^*$};
\end{pgfonlayer}
\end{tikzpicture}
}
\end{subfigure}%
\begin{subfigure}{0.3\linewidth}
\scalebox{0.67}{
\begin{tikzpicture}[cirtikz, minimum height=45pt]
    \node (i1) [fill=yellow3, anchor=north west] {\Gaussian};
    \node (i2) [fill=tomato3, anchor=north, below=15pt of i1.south] {\Gaussian};
    \node (p1) [fill=lacamgold4, anchor=west, right=20pt of i1.east] at ($(i1.east)!0.5!(i2.east)$) {\Odot};
    \node (r1) [fill=lacamgold4, anchor=west, right=20pt of p1.east] {\Sum};
    \node (d2) [fill=petroil4, minimum width=45pt, below=20pt of r1.south east, anchor=north east] {\color{white} \LARGE $\bm{I}_K$};
    \node (d1) [fill=lacamgold4, minimum width=45pt, minimum height=45pt,  anchor=west, right=20pt of d2.east] {\color{white} \LARGE $\bm{\vo \otimes \vo^*}$};
    \node (q) [fill=olive4, minimum width=45pt, anchor=north, below=20pt of d1.south] at ($(d2.south)!0.5!(d1.south)$) {\Odot};
    \node (s) [fill=olive4, minimum width=20pt, minimum height=20pt, anchor=north, below=20pt of q.south] {\Sum};
\begin{pgfonlayer}{background}
    \node (non-marg-bg) [ultra thick, draw=tomato3!40!bgrey3, dotted, anchor=north west, minimum width=112pt, minimum height=112pt, outer sep=-5pt] {};
    \draw[bcedge, inner sep=-2pt,-{Stealth}, shorten >=3pt, shorten <=3pt] (r1.east) -| (d1.north) node[midway, above right=-1em and -1.5em] {\LARGE \color{black} $\vo\in\bbC^K$};
    \node (marg-bg) [ultra thick, draw=deliveroo!40!bgrey3, dotted, minimum width=56pt, minimum height=56pt, outer sep=-5pt] at (d2) {};
    \node (mixed-bg) [ultra thick, draw=olive4!60!bgrey3, dotted, above=1pt of q.north, anchor=north, minimum width=56pt, minimum height=94pt, outer sep=-5pt] {};
    \draw [-, bcedge] (i1.south east) -- (i2.north east);
    \draw [-, bcedge] (i1.north east) -- (p1.north west);
    \draw [-, bcedge] (i2.south east) -- (p1.south west);
    \draw [-, bcedge] (p1.north east) -- (r1.north west);
    \draw [-, bcedge] (p1.south east) -- (r1.south west);
    \draw [-, bcedge] (d1.south west) -- (d2.south east);
    \draw [-, bcedge] (d1.south east) -- (q.north east);
    \draw [-, bcedge] (d2.south west) -- (q.north west);
    \draw [-, bcedge] (q.south west) -- (s.north west);
    \draw [-, bcedge] (q.south east) -- (s.north east);
    \draw [o-, bcedge, fill=bdark, shorten <=-4pt, shorten >=-15pt] ($(p1.east)!0.5!(r1.west)$) -- ($(p1.north east)!0.5!(r1.north west)$) node [yshift=22pt] {\large\color{black} $\vW_2$};
    \draw [o-, bcedge, fill=bdark, shorten <=-4pt, shorten >=-15pt] ($(q.south)!0.5!(s.north)$) -- ($(q.south west)!0.5!(s.north west)$) node [xshift=-44pt] {\large\color{black} $\vW_1\otimes\vW_1^*$};
\end{pgfonlayer}
\begin{pgfonlayer}{foreground}
    \node (i1l) [left=8pt of i1.west] {\large $f_1(X_1)$};
    \node (i2l) [left=8pt of i2.west] {\large $f_2(X_2)$};
    \node (phi-evidence) [text height=6pt, text width=10pt, anchor=south west, right=88pt of i1.east] {\LARGE $\phi_{\vY}$};
    \node (phi-marginalized) [text height=6pt, text width=10pt, anchor=south east, left=10pt of marg-bg.south west] {\LARGE $\phi_{\vZ}$};
    \node (phi-mixed) [text height=6pt, text width=10pt, anchor=west, right=18pt of s.east] {\LARGE $\phi_{\vY,\vZ}$};
\end{pgfonlayer}
\end{tikzpicture}
}
\vspace{-.5em}
\end{subfigure}%
\vspace{-.5em}
    \caption{\textbf{Squared orthonormal PCs 
    enable a more efficient marginalization algorithm.} %
    The \textbf{left} figure shows a tensorized circuit $c$ with a tree structure over $\vX = \{X_1,X_2,X_3,X_4\}$ with input \scalebox{0.45}{\protect\begin{tikzpicture} \protect\node [fill=bgrey3, minimum width=45pt] {\protect\Gaussian}; \protect\end{tikzpicture}}, Hadamard \scalebox{0.45}{\protect\begin{tikzpicture}\protect \node [fill=bgrey3, minimum width=45pt] {\protect\Odot}; \protect\end{tikzpicture}} and sum \scalebox{0.45}{\protect \begin{tikzpicture} \protect\node [fill=bgrey3, minimum width=45pt] {\protect\Sum}; \protect\end{tikzpicture}} layers.
    We label the input layers with the vector-valued function they encode on a variable $X_i$.
    Consider computing the marginal likelihood $p(x_1,x_2) = \int_{\dom(X_3)\times \dom(X_4)} |c(x_1,x_2,x_3,x_4)|^2 \mathrm{d}x_3\mathrm{d}x_4$.
    We label group of layers depending on $\vY=\{X_1,X_2\}$ (red-ish, $\phi_{\vY}$), $\vZ=\{X_3,X_4\}$ (blue-ish, $\phi_{\vZ}$), and on both (green, $\phi_{\vY,\vZ}$).
    A naive algorithm computing $p(x_1,x_2)$ would (i) square the \emph{whole} tensorized circuit as 
    $c^2$, where the size of each layer quadratically increases, and (ii) compute the integrals of squared input layers over $\vZ$ and (iii) evaluate the rest of the squared layers (\textbf{middle}, from left to right).
    (\textbf{right}) Instead, if $c$ is orthonormal, \cref{alg:marginalization} avoids the computation of the integral of the sub-circuit depending on $\vZ$ (as it results in the identity matrix $\vI_K$, in blue), and requires computing a single Kronecker product (orange) and squaring just the layers in $\phi_{\vY,\vZ}$ (green).
    }
    \label{fig:tighter-marginalization}
\end{figure*}

\vspace{-.5em}

\section{Squared Orthonormal Circuits}\label{sec:normalized-squared}

Our representation of squared PCs is based on the definition of orthonormal circuits we introduce below.

\begin{defn}[Orthonormal circuits]
    \label{defn:orthonormal-circuit}
    A tensorized circuit $c$ over variables $\vX$ is said \emph{orthonormal} if
    (1) each input layer $\vell$ over $X\in\vX$ encodes a vector of $K$ orthonormal functions, i.e., $\vell(X) = [f_1(X),\ldots,f_K(X)]^\top$ such that $\forall i,j\in[K]^2\colon \int_{\dom(X)} f_i(x)f_j(x)^* \mathrm{d}x = \delta_{ij}$, where $\delta_{ij}$ denotes the Kronecker delta; and
    (2) each sum layer is parameterized by a (semi-)unitary matrix $\vW\in\bbC^{K_1\times K_2}$, $K_1\leq K_2$, i.e., $\vW\vW^\dagger = \vI_{K_1}$, or the rows of $\vW$ are orthonormal.
\end{defn}

If we take a tensorized circuit $c$ that is orthonormal, then the squared PC obtained from $c$ is guaranteed to encode a normalized probability distribution, as formalized below.

\begin{prop}
    \label{prop:orthonormal-circuits-normalization}
    Let $c$ be a structured-decomposable tensorized circuit over variables $\vX$.
    If $c$ is orthonormal, then its squaring encodes a normalized distribution, i.e., $Z=1$.
\end{prop}

\cref{app:normalized-squared} shows our proof.
The idea is that integrating products of input layers encoding orthonormal functions yields identity matrices in $c^2$.
Then, the (semi-)unitarity of parameter matrices in sum layers is used to show the output of each layer in $c^2$ is (the flattening of) an identity matrix, thus eventually yielding $Z = 1$.

The computation of the partition function $Z$ represents a special case of marginalization, where all variables are marginalized out.
In general, computing any marginal probabilities in squared PCs requires worst-case time $\calO(LS^2)$ \citep{loconte2024subtractive}.
In the next section, we show how to exploit the properties of orthonormal circuits (\cref{defn:orthonormal-circuit}) to also provide an algorithm that computes any marginal probability with a better complexity.

\section{A Tighter Marginalization Complexity}\label{sec:tighter-marginalization}

The idea of our algorithm is that, when computing marginal probabilities using $c^2$, \emph{we do not need to evaluate the layers whose scope depends on only the variables being integrated out.}
This is because they would always result in identity matrices, as noticed in our proof for \cref{prop:orthonormal-circuits-normalization}.

In addition, we observe that we do not need to square the \emph{whole} tensorized circuit $c$, \emph{but only a fraction of the layers depending on} both \emph{the marginalized variables and the ones being kept}.
By doing so, a part of the complexity will depend on $S$ rather than $S^2$.
We formalize our result below.

\begin{thm}
    \label{thm:marginalization-complexity}
    Let $c$ be a structured-decomposable orthonormal circuit over variables $\vX$.
    Let $\vZ\subseteq \vX$, $\vY=\vX\setminus\vZ$.
    Computing the marginal likelihood $p(\vy) = \int_{\dom(\vZ)} |c(\vy,\vz)|^2 \mathrm{d}\vz$ requires time $\calO(|\phi_{\vY}| S + |\phi_{\vY,\vZ}| S^2)$, where $\phi_{\vY}$ (resp. $\phi_{\vY,\vZ}$) denotes the set of layers in $c$ whose scope depends on only variables in $\vY$ (resp. on variables both in $\vY$ \emph{and} in $\vZ$).
\end{thm}

We prove it in \cref{app:marginalization} and show our algorithm in \cref{alg:marginalization}.
Note that the complexity shown in \cref{thm:marginalization-complexity} is independent on the number of layers whose scope depend on $\vZ$ only, i.e., $\phi_{\vZ}$.
Depending on the circuit structure and the variables $\vZ$, \cref{alg:marginalization} can be much more efficient than $\calO(LS^2)$.
For example, the tree structure of a circuit defined over pixel variables can be built by recursively splitting an image into patches with horizontal and vertical cuts \citep{loconte2024relationship}.
If $\vZ$ consists of only the pixel variables of the left-hand side of an image (i.e., we are computing the marginal of the right-hand side $\vY$), then $|\phi_{\vY,\vZ}|\ll L$ since only a few layers near the root will depend on both $\vY$ and $\vZ$.
We illustrate an example in \cref{fig:tighter-marginalization}.

\section{Are Orthonormal Circuits less Expressive?}\label{sec:question-expressiveness}

Orthonormal tensorized circuits restrict their input layers to encode orthonormal functions, and require parameter matrices to be (semi-)unitary (\cref{defn:orthonormal-circuit}), thus arising the question whether these conditions make them less expressive when compared to non-orthonormal ones.
Below, we start by analyzing which input layer functions we can encode in terms of orthonormal functions.

\paragraph{Are orthonormal functions restrictive?}
Depending on whether a variable is discrete or continuous, we have different ways to model it with orthonormal functions.
For a discrete variable $X$ with finite domain $\dom(X) = [v]$, any function $f(X)$ can be expressed as $f(x) = \sum_{k=1}^v f(k)\delta_{xk}$, i.e., $f$ can be written in terms of $v$ basis functions $\{\delta_{xk}\}_{k=1}^v$ that are orthonormal.
That is, we have that $\sum_{x\in\dom(X)} \delta_{xk}\delta_{xk'} = \delta_{kk'}$ for $k,k'\in[v]$.
Therefore, any input layer over a finitely-discrete variable $X$ can be exactly encoded with a sum layer having a layer encoding the orthonormal basis $\{\delta_{xk}\}_{k=1}^v$ as input.

In the case of a continuous variable $X$, many classes of functions can be expressed in terms of orthonormal basis functions.
For instance, periodic functions can be represented by Fourier series of sines and cosines basis that form an orthonormal set of functions \citep{jackson1941fourier}.
Under certain continuity conditions, functions can be approximated arbitrarily well by finite Fourier partial sums \citep{jackson1982theory}.
Moreover, depending on the support of $X$, many classes of functions can be described in terms of families of orthonormal polynomials, such as Legendre, Laguerre and Hermite polynomials \citep{abramowitz1965handbook}.
Notably, Hermite functions generalize Gaussians and are a basis of square-integrable functions over all $\bbR$ \citep{roman1984umbral}.

\paragraph{Are (semi-)unitary matrices restrictive?}
Next, we investigate whether the requirement of sum layer parameter matrices to be (semi-)unitary may reduce the expressiveness of squared PCs.
In the following, we answer to this question negatively, as we can enforce this condition in polynomial time w.r.t the number of layers and the layer size.

\begin{thm}
    \label{thm:orthogonalization-algorithm}
    Let $c$ be a tensorized circuit over variables $\vX$.
    Assume that each input layer in $c$ encodes a set of orthonormal functions.
    Then, there exists an algorithm returning an orthonormal circuit $c'$ in polynomial time such that $c'$ is equivalent to $c$ up to a multiplicative constant, i.e., $c'(\vX) = Z^{-\frac{1}{2}} c(\vX)$ where $Z = \int_{\dom(\vX)} |c(\vx)|^2 \mathrm{d}\vx$.
\end{thm}

\cref{app:question-expressiveness} presents our proof, and \cref{alg:orthonormalization} shows our algorithm to ``\emph{orthonormalize}'' a tensorized circuit.
The idea of \cref{alg:orthonormalization} is that we can recursively make sub-circuits orthonormal by (i) factorizing the sum layer parameter matrices via QR decompositions, and (ii) by ``\emph{pushing}'' the non-unitary part of such a decomposition towards the output, until the reciprocal square root of the partition function of $c^2$ is retrieved at the top level of the recursion.
\cref{fig:orthonormalization-transformation} illustrates the algorithm.
Therefore, \cref{thm:orthogonalization-algorithm} guarantees that squared orthonormal PCs are as expressive as general squared PCs with orthonormal input functions.

Finally, we note that \cref{alg:orthonormalization} is the dual of a result about monotonic PCs shown by \citet{peharz2015theoretical}: they show an algorithm that updates the positive weights of a smooth and decomposable PC such that the distribution it encodes is already normalized.
Here, we show an algorithm that updates the (possibly) complex weights of a structured-decomposable circuit such that its squaring encodes an already-normalized probability distribution.

\begin{algorithm}[!t]
\small
\caption{$\textsc{Orthonormalize}(\vell)$}\label{alg:orthonormalization}
    \textbf{Input:} The output layer $\vell$ of a structured-decomposable circuit, whose input layers encode a set of orthonormal functions.\\
    \textbf{Output:} The output layer of a structured-decomposable circuit $c'$ that is orthonormal, and a matrix $\vR\in\bbC^{K_1\times K_2}$, $K_1\leq K_2$.
\begin{algorithmic}[1]
    \If{$\vell$ is an input layer}
        \State Assume $\vell$ outputs vectors in $\bbC^K$
        \State \Return $(\vell, \vI_K)$
    \EndIf
    \If{$\vell$ is a sum layer with input $\vell_1$ and parameterized by $\vW$}
        \State $(\vell_1', \vR_1) \leftarrow \textsc{Orthonormalize}(\vell_1)$, $\vR_1\in\bbC^{K_2\times K_3}$
        \State Assume $\vW\in\bbC^{K_1\times K_2}$, $K_1\leq K_2$
        \State Let $\vV = \vW\vR_1 \in\bbC^{K_1\times K_3}$, $K_1\leq K_3$
        \State Factorize $\vV^\dagger = \vQ\vR$ where $\vQ\in\bbC^{K_3\times K_1}$, $\vR\in\bbC^{K_1\times K_1}$
        \State \qquad such that $\vQ^\dagger\vQ = \vI_{K_1}$ and $\vR$ is upper triangular
        \State Let $\vell'$ be a sum layer computing $\vQ^\dagger \vell_1'(\uscope(\vell))$
        \State \Return $(\vell',\vR^\dagger)$
    \EndIf
    \If{$\vell$ is a Kronecker product layer with inputs $\vell_1,\vell_2$}
        \State $(\vell_1',\vR_1)\leftarrow\textsc{Orthonormalize}(\vell_1)$, $\vR_1\in\bbC^{K_1\times K_2}$
        \State $(\vell_2',\vR_2)\leftarrow\textsc{Orthonormalize}(\vell_2)$, $\vR_2\in\bbC^{K_3\times K_4}$
        \State Let $\vell'$ be a layer computing $\vell_1'(\uscope(\vell_1))\otimes \vell_2'(\uscope(\vell_2))$
        \State \Return $(\vell',\vR_1\otimes\vR_2)$
        \Comment{$\otimes$: Kronecker matrix product}
    \EndIf
    \If{$\vell$ is an Hadamard product layer with inputs $\vell_1,\vell_2$}
        \State $(\vell_1',\vR_1)\leftarrow\textsc{Orthonormalize}(\vell_1)$, $\vR_1\in\bbC^{K_1\times K_2}$
        \State $(\vell_2',\vR_2)\leftarrow\textsc{Orthonormalize}(\vell_2)$, $\vR_2\in\bbC^{K_1\times K_3}$
        \State Let $\vell'$ be a layer computing $\vell_1'(\uscope(\vell_1))\otimes \vell_2'(\uscope(\vell_2))$
        \State \Return $(\vell',\vR_1\bullet\vR_2)$
        \Comment{$\bullet$: Face-splitting matrix product}
    \EndIf
\end{algorithmic}
\end{algorithm}

\section{Related Work and Conclusion}\label{sec:conclusion}

In this paper, we presented a parameterization of squared PCs inspired by canonical forms in TNs, based on orthonormal functions and (semi-)unitary matrices, as to speed-up the computation of marginal probabilities.
As squared orthonormal PCs support faster marginalization, they are amenable for future works on applications where computing marginals is key, e.g., lossless compression \citep{yang2022neural,liu2022lossless}.

Orthonormal basis functions have been used to parameterize \emph{shallow} squared mixtures encoding already-normalized distributions, both in signal processing \citep{pinheiro1997estimating} and in score-based variational inference \citep{cai2024eigenvi}.
Our squared orthonormal PCs generalize them, as they represent \emph{deeper} squared mixtures.

We plan to investigate different methods as to learn squared orthonormal PCs from data for distribution estimation, and compare how do they perform w.r.t. squared PCs with unconstrained parameters.
For instance, there are many ways of parameterizing unitary matrices, with different advantages regarding efficiency, numerical stability, and optimization \citep{arjovsky2015unitary,huang2017orthogonal,bansal2018can-orthogonality,casado2019cheap-orthogonal}.
Moreover, \citet{hauru2020riemannian,luchnikov2021qgopt} proposed optimizing the parameters of MPS TNs and quantum gates by performing gradient descent over the Stiefel manifold \citep{absil2007optimization}.
Furthermore, recent works parameterize more expressive monotonic PCs using neural networks \citep{shao2022conditional,gala2024pic,gala2024scaling}, thus motivating parameterizing squared orthonormal PCs similarly.

\section*{Acknowledgments}
We acknowledge Raul Garcia-Patron Sanchez for meaningful discussions about tensor networks and quantum circuits.
AV was supported by the ``UNREAL: Unified Reasoning Layer for Trustworthy ML'' project (EP/Y023838/1) selected by the ERC and funded by UKRI EPSRC.

\section*{Contributions}
LL and AV conceived the initial idea of the paper.
LL is responsible for all theoretical contributions, pictures, algorithms and writing.
AV supervised all the phases of the project and provided feedback.

\bibliography{referomnia}

\begin{thebibliography}{41}
\providecommand{\natexlab}[1]{#1}

\bibitem[{Abramowitz, Stegun, and Miller(1965)}]{abramowitz1965handbook}
Abramowitz, M.; Stegun, I.~A.; and Miller, D. 1965.
\newblock Handbook of Mathematical Functions With Formulas, Graphs and
  Mathematical Tables (National Bureau of Standards Applied Mathematics Series
  No. 55).
\newblock \emph{Journal of Applied Mechanics}, 32: 239--239.

\bibitem[{Absil, Mahony, and Sepulchre(2007)}]{absil2007optimization}
Absil, P.-A.; Mahony, R.~E.; and Sepulchre, R. 2007.
\newblock \emph{Optimization Algorithms on Matrix Manifolds}.
\newblock Princeton University Press.

\bibitem[{Ahmed et~al.(2022)Ahmed, Teso, Chang, Van~den Broeck, and
  Vergari}]{ahmed2022semantic}
Ahmed, K.; Teso, S.; Chang, K.-W.; Van~den Broeck, G.; and Vergari, A. 2022.
\newblock Semantic probabilistic layers for neuro-symbolic learning.
\newblock In \emph{Advances in Neural Information Processing Systems 35
  (NeurIPS)}, volume~35, 29944--29959. Curran Associates, Inc.

\bibitem[{Arjovsky, Shah, and Bengio(2015)}]{arjovsky2015unitary}
Arjovsky, M.; Shah, A.; and Bengio, Y. 2015.
\newblock Unitary Evolution Recurrent Neural Networks.
\newblock In \emph{International Conference on Machine Learning}.

\bibitem[{Bansal, Chen, and Wang(2018)}]{bansal2018can-orthogonality}
Bansal, N.; Chen, X.; and Wang, Z. 2018.
\newblock Can We Gain More from Orthogonality Regularizations in Training Deep
  Networks?
\newblock In \emph{Advances in Neural Information Processing Systems},
  volume~31.

\bibitem[{Biamonte and Bergholm(2017)}]{biamonte2017tensor}
Biamonte, J.~D.; and Bergholm, V. 2017.
\newblock Tensor Networks in a Nutshell.
\newblock \emph{arXiv: Quantum Physics}.

\bibitem[{Cai et~al.(2024)Cai, Modi, Margossian, Gower, Blei, and
  Saul}]{cai2024eigenvi}
Cai, D.; Modi, C.; Margossian, C.; Gower, R.~M.; Blei, D.; and Saul, L.~K.
  2024.
\newblock Eigen{VI}: score-based variational inference with orthogonal function
  expansions.
\newblock In \emph{The Thirty-eighth Annual Conference on Neural Information
  Processing Systems ({NeurIPS})}.

\bibitem[{Casado and Mart{\'i}nez-Rubio(2019)}]{casado2019cheap-orthogonal}
Casado, M.~L.; and Mart{\'i}nez-Rubio, D. 2019.
\newblock Cheap Orthogonal Constraints in Neural Networks: A Simple
  Parametrization of the Orthogonal and Unitary Group.
\newblock \emph{ArXiv}, abs/1901.08428.

\bibitem[{Cheng et~al.(2019)Cheng, Wang, Xiang, and Zhang}]{cheng2019tree}
Cheng, S.; Wang, L.; Xiang, T.; and Zhang, P. 2019.
\newblock Tree tensor networks for generative modeling.
\newblock \emph{Physical Review B}, 99(15): 155131.

\bibitem[{Choi, Vergari, and Van~den Broeck(2020)}]{choi2020pc}
Choi, Y.; Vergari, A.; and Van~den Broeck, G. 2020.
\newblock Probabilistic Circuits: A Unifying Framework for Tractable
  Probabilistic Modeling.
\newblock Technical report, University of California, Los Angeles (UCLA).

\bibitem[{Darwiche and Marquis(2002)}]{darwiche2002knowledge}
Darwiche, A.; and Marquis, P. 2002.
\newblock A knowledge compilation map.
\newblock \emph{Journal of Artificial Intelligence Research (JAIR)}, 17:
  229--264.

\bibitem[{Gala et~al.(2024{\natexlab{a}})Gala, de~Campos, Peharz, Vergari, and
  Quaeghebeur}]{gala2024pic}
Gala, G.; de~Campos, C.; Peharz, R.; Vergari, A.; and Quaeghebeur, E.
  2024{\natexlab{a}}.
\newblock Probabilistic Integral Circuits.
\newblock In \emph{AISTATS 2024}.

\bibitem[{Gala et~al.(2024{\natexlab{b}})Gala, de~Campos, Vergari, and
  Quaeghebeur}]{gala2024scaling}
Gala, G.; de~Campos, C.; Vergari, A.; and Quaeghebeur, E. 2024{\natexlab{b}}.
\newblock Scaling Continuous Latent Variable Models as Probabilistic Integral
  Circuits.
\newblock \emph{arXiv preprint arXiv:2406.06494}.

\bibitem[{Glasser, Pancotti, and Cirac(2018)}]{glasser2018from}
Glasser, I.; Pancotti, N.; and Cirac, J.~I. 2018.
\newblock From Probabilistic Graphical Models to Generalized Tensor Networks
  for Supervised Learning.
\newblock \emph{IEEE Access}, 8: 68169--68182.

\bibitem[{Glasser et~al.(2019)Glasser, Sweke, Pancotti, Eisert, and
  Cirac}]{glasser2019expressive}
Glasser, I.; Sweke, R.; Pancotti, N.; Eisert, J.; and Cirac, I. 2019.
\newblock Expressive power of tensor-network factorizations for probabilistic
  modeling.
\newblock In \emph{Advances in Neural Information Processing Systems 32
  (NeurIPS)}, 1498--1510. Curran Associates, Inc.

\bibitem[{Hauru, Damme, and Haegeman(2020)}]{hauru2020riemannian}
Hauru, M.; Damme, M.~V.; and Haegeman, J. 2020.
\newblock Riemannian optimization of isometric tensor networks.
\newblock \emph{SciPost Physics}.

\bibitem[{Huang et~al.(2017)Huang, Liu, Lang, Yu, and Li}]{huang2017orthogonal}
Huang, L.; Liu, X.; Lang, B.; Yu, A.~W.; and Li, B. 2017.
\newblock Orthogonal Weight Normalization: Solution to Optimization over
  Multiple Dependent Stiefel Manifolds in Deep Neural Networks.
\newblock In \emph{AAAI Conference on Artificial Intelligence}.

\bibitem[{Jackson(1941)}]{jackson1941fourier}
Jackson, D. 1941.
\newblock Fourier series and orthogonal polynomials.

\bibitem[{Jackson(1982)}]{jackson1982theory}
Jackson, D. 1982.
\newblock The theory of approximation.

\bibitem[{Liu, Mandt, and Van~den Broeck(2022)}]{liu2022lossless}
Liu, A.; Mandt, S.; and Van~den Broeck, G. 2022.
\newblock Lossless Compression with Probabilistic Circuits.
\newblock In \emph{International Conference on Learning Representations}.

\bibitem[{Loconte et~al.(2024{\natexlab{a}})Loconte, Aleksanteri, Mengel,
  Trapp, Solin, Gillis, and Vergari}]{loconte2024subtractive}
Loconte, L.; Aleksanteri, M.~S.; Mengel, S.; Trapp, M.; Solin, A.; Gillis, N.;
  and Vergari, A. 2024{\natexlab{a}}.
\newblock Subtractive Mixture Models via Squaring: Representation and Learning.
\newblock In \emph{The Twelfth International Conference on Learning
  Representations ({ICLR})}.

\bibitem[{Loconte et~al.(2023)Loconte, {Di Mauro}, Peharz, and
  Vergari}]{loconte2023turn}
Loconte, L.; {Di Mauro}, N.; Peharz, R.; and Vergari, A. 2023.
\newblock How to Turn Your Knowledge Graph Embeddings into Generative Models
  via Probabilistic Circuits.
\newblock In \emph{Advances in Neural Information Processing Systems 37
  (NeurIPS)}. Curran Associates, Inc.

\bibitem[{Loconte et~al.(2024{\natexlab{b}})Loconte, Mari, Gala, Peharz,
  de~Campos, Quaeghebeur, Vessio, and Vergari}]{loconte2024relationship}
Loconte, L.; Mari, A.; Gala, G.; Peharz, R.; de~Campos, C.; Quaeghebeur, E.;
  Vessio, G.; and Vergari, A. 2024{\natexlab{b}}.
\newblock What is the Relationship between Tensor Factorizations and Circuits
  (and How Can We Exploit it)?
\newblock arXiv:2409.07953.

\bibitem[{Loconte, Mengel, and Vergari(2024)}]{loconte2024sos}
Loconte, L.; Mengel, S.; and Vergari, A. 2024.
\newblock Sum of Squares Circuits.
\newblock arXiv:2408.11778.

\bibitem[{Luchnikov et~al.(2021)Luchnikov, Ryzhov, Filippov, and
  Ouerdane}]{luchnikov2021qgopt}
Luchnikov, I.~A.; Ryzhov, A.; Filippov, S.~N.; and Ouerdane, H. 2021.
\newblock QGOpt: Riemannian optimization for quantum technologies.
\newblock \emph{SciPost Physics}.

\bibitem[{Novikov, Panov, and Oseledets(2021)}]{novikov2021ttde}
Novikov, G.~S.; Panov, M.~E.; and Oseledets, I.~V. 2021.
\newblock Tensor-train density estimation.
\newblock In \emph{37th Conference on Uncertainty in Artificial Intelligence
  (UAI)}, volume 161 of \emph{Proceedings of Machine Learning Research},
  1321--1331. PMLR.

\bibitem[{Or{\'u}s(2013)}]{orus2013practical}
Or{\'u}s, R. 2013.
\newblock A Practical Introduction to Tensor Networks: Matrix Product States
  and Projected Entangled Pair States.
\newblock \emph{Annals of Physics}, 349: 117--158.

\bibitem[{Peharz et~al.(2015)Peharz, Tschiatschek, Pernkopf, and
  Domingos}]{peharz2015theoretical}
Peharz, R.; Tschiatschek, S.; Pernkopf, F.; and Domingos, P.~M. 2015.
\newblock {On Theoretical Properties of Sum-Product Networks}.
\newblock In \emph{International Conference on Artificial Intelligence and
  Statistics}.

\bibitem[{P{\'e}rez-Garc{\'i}a et~al.(2007)P{\'e}rez-Garc{\'i}a, Verstraete,
  Wolf, and Cirac}]{perez2006mps}
P{\'e}rez-Garc{\'i}a, D.; Verstraete, F.; Wolf, M.~M.; and Cirac, J.~I. 2007.
\newblock Matrix Product State Representations.
\newblock \emph{Quantum Information and Computing}, 7(5): 401–430.

\bibitem[{Pinheiro and Vidakovic(1997)}]{pinheiro1997estimating}
Pinheiro, A.; and Vidakovic, B. 1997.
\newblock Estimating the square root of a density via compactly supported
  wavelets.
\newblock \emph{Computational Statistics and Data Analysis}, 25(4): 399--415.

\bibitem[{Pipatsrisawat and Darwiche(2008)}]{pipatsrisawat2008new}
Pipatsrisawat, K.; and Darwiche, A. 2008.
\newblock New Compilation Languages Based on Structured Decomposability.
\newblock In \emph{23rd Conference on Artificial Intelligence (AAAI)},
  volume~8, 517--522.

\bibitem[{Roman(1984)}]{roman1984umbral}
Roman, S. 1984.
\newblock The Umbral Calculus.

\bibitem[{Schollwoeck(2010)}]{schollwoeck2010density}
Schollwoeck, U. 2010.
\newblock The density-matrix renormalization group in the age of matrix product
  states.
\newblock \emph{Annals of Physics}, 326: 96--192.

\bibitem[{Shao et~al.(2022)Shao, Molina, Vergari, Stelzner, Peharz, Liebig, and
  Kersting}]{shao2022conditional}
Shao, X.; Molina, A.; Vergari, A.; Stelzner, K.; Peharz, R.; Liebig, T.; and
  Kersting, K. 2022.
\newblock Conditional sum-product networks: Modular probabilistic circuits via
  gate functions.
\newblock \emph{International Journal of Approximate Reasoning}, 140: 298--313.

\bibitem[{Shi, Duan, and Vidal(2006)}]{shi2006tree}
Shi, Y.-Y.; Duan, L.-M.; and Vidal, G. 2006.
\newblock Classical simulation of quantum many-body systems with a tree tensor
  network.
\newblock \emph{Physical Review A}, 74: 22320.

\bibitem[{Shpilka and Yehudayoff(2010)}]{shpilka2010open}
Shpilka, A.; and Yehudayoff, A. 2010.
\newblock Arithmetic Circuits: A survey of recent results and open questions.
\newblock \emph{Founddations and Trends in Theoretical Computer Science}, 5:
  207--388.

\bibitem[{Stoudenmire and Schwab(2016)}]{stoudenmire2016supervised}
Stoudenmire, E.; and Schwab, D.~J. 2016.
\newblock Supervised Learning with Tensor Networks.
\newblock In \emph{Advances in Neural Information Processing Systems 29
  (NeurIPS)}, 4799--4807. Curran Associates, Inc.

\bibitem[{Vergari et~al.(2021)Vergari, Choi, Liu, Teso, and Van~den
  Broeck}]{vergari2021compositional}
Vergari, A.; Choi, Y.; Liu, A.; Teso, S.; and Van~den Broeck, G. 2021.
\newblock A Compositional Atlas of Tractable Circuit Operations for
  Probabilistic Inference.
\newblock In \emph{Advances in Neural Information Processing Systems 34
  (NeurIPS)}, 13189--13201. Curran Associates, Inc.

\bibitem[{Vergari, {Di Mauro}, and Esposito(2019)}]{vergari2019visualizing}
Vergari, A.; {Di Mauro}, N.; and Esposito, F. 2019.
\newblock Visualizing and understanding sum-product networks.
\newblock \emph{Machine Learning}, 108(4): 551--573.

\bibitem[{Yang, Mandt, and Theis(2022)}]{yang2022neural}
Yang, Y.; Mandt, S.; and Theis, L. 2022.
\newblock An Introduction to Neural Data Compression.
\newblock \emph{Foundations and Trends in Computer Graphics and Vision}, 15:
  113--200.

\bibitem[{Zhang et~al.(2023)Zhang, Dang, Peng, and Van~den
  Broeck}]{zhang2023tractable}
Zhang, H.; Dang, M.; Peng, N.; and Van~den Broeck, G. 2023.
\newblock Tractable Control for Autoregressive Language Generation.
\newblock In \emph{40th International Conference on Machine Learning (ICML)},
  volume 202 of \emph{Proceedings of Machine Learning Research}, 40932--40945.
  {PMLR}.

\end{thebibliography}

\clearpage
\appendix

\counterwithin{table}{section}
\counterwithin{figure}{section}
\counterwithin{algorithm}{section}
\renewcommand{\thetable}{\thesection.\arabic{table}}
\renewcommand{\thefigure}{\thesection.\arabic{figure}}
\renewcommand{\thealgorithm}{\thesection.\arabic{algorithm}}

\section{Squaring Tensorized Circuits}\label{app:squaring-circuits}

\begin{algorithm}[!t]
    \caption{$\textsc{SquareTensorizedCircuit}(\vell)$}\label{alg:tensorized-square}
    \textbf{Input:} A tensorized circuit with output layer $\vell$ that is structured-decomposable.\\
    \textbf{Output:} The tensorized squared circuit having $\vell^2$ as output layer computing $\vell\otimes \vell^*$.
    \begin{algorithmic}[1]
        \If{$\vell$ is an input layer}
            \State $\vell$ computes $K$ functions $\{f_i\}_{i=1}^K$ over $X$
            \State \Return An input layer $\vell^2$ computing all $K^2$\\
            \qquad\qquad product combinations $f_i(X)f_j^*(X)$
        \EndIf
        \If{$\vell$ is a product layer}
            \State $(\vell_1,\vell_2) \leftarrow \textsc{GetInputs}(\vell)$
            \State $\vell_1^2 \leftarrow \textsc{SquareTensorizedCircuit}(\vell_1)$
            \State $\vell_2^2 \leftarrow \textsc{SquareTensorizedCircuit}(\vell_2)$
            \If{$\vell$ is an Hadamard product layer}
                \State \Return $\vell_1^2(\uscope(\vell_1)) \odot \vell_2^2(\uscope(\vell_2))$
            \Else
                \Comment{$\vell$ is Kronecker product layer}
                \State \Return $\vP \left( \vell_1^2(\uscope(\vell_1)) \otimes \vell_2^2(\uscope(\vell_2)) \right)$,\\
                \qquad\qquad\qquad where $\vP$ is a permutation matrix
            \EndIf
        \EndIf
        \Comment{$\vell$ is a sum layer}
        \State $(\vell_1,) \leftarrow \textsc{GetInputs}(\vell)$
        \State $\vell_1^2 \leftarrow \textsc{SquareTensorizedCircuit}(\vell_1)$
        \State $\vW \in \bbR^{S\times K} \leftarrow \textsc{GetParameters}(\vell)$
        \State $\vW' \in \bbR^{S^2\times K^2} \leftarrow \vW \otimes \vW^*$
        \State \Return $\vW' \vell_1^2(\uscope(\vell_1))$
    \end{algorithmic}
\end{algorithm}

Given a tensorized circuit $c$, a squared PC models $p(\vX) = Z^{-1}|c(\vX)|^2 = Z^{-1} c(\vX)c(\vX)^*$.
To compute $Z$ efficiently, one has to represent $|c(\vX)|^2$ as a decomposable circuit (\cref{defn:smoothness-decomposability}), since it would allow tractable variable marginalization \citep{choi2020pc}.

\cref{alg:tensorized-square} recursively constructs the circuit $c^2$ computing $|c(\vX)|^2$ as yet another decomposable tensorized circuit.
\cref{alg:tensorized-square} is taken from \citet{loconte2024subtractive}, but here we trivially generalize it as to allow complex weight parameter matrices.
In this algorithm, each layer $\vell$ in $c$ is recursively squared into a layer $\vell^2$ in $c^2$ as to compute the Kronecker product between the output of $\vell$ and its conjugate.
As a consequence, the size of each layer in $c$ is quadratically increased in $c^2$.
For instance, each input layer $\vell$ in $c$ over a variable $X$ and computing $\vell(X)\in\bbC^K$ is squared as to recover an input layer $\vell^2$ in $c^2$ such that $\vell^2(X) = \vell(X)\otimes \vell(X)^* \in\bbC^{K^2}$ (L1-4).
An Hadamard (resp. Kronecker) product layer in $c$ is squared as another Hadamard layer (resp. a composition of sum and Kronecker layers) in $c^2$ (L5-13).
Finally, a sum layer $\vell$ in $c$ computing $\vW\vell_1(\uscope(\vell_1))$ is squared as to recover a sum layer $\vell^2$ in $c^2$ that is parameterized by $\vW\otimes\vW^*$ instead (L14-18).
\cref{fig:tighter-marginalization} (left) shows an example of tensorized circuit and \cref{fig:tighter-marginalization} (middle) shows its squaring obtained with \cref{alg:tensorized-square}.

\section{Proofs}\label{app:proofs}

We start with some notation.

\paragraph{Notation.} In \cref{prop:orthonormal-circuits-normalization} and \cref{thm:marginalization-complexity} we require integrating layers $\vell$ that output vectors, e.g., in $\bbC^K$.
That is, given a layer $\vell$ having scope $\uscope(\vell) = \vY\cup\vZ$ and encoding a function $\vell\colon\dom(\vY\cup\vZ)\to\bbC^K$, we write $\int_{\dom(\vZ)} \vell(\vy,\vz) \mathrm{d}\vz$ to refer to the $K$-dimensional vector obtained by integrating the $K$ function components encoded by $\vell$, i.e., $\int_{\dom(\vZ)} \vell(\vy,\vz) \mathrm{d}\vz =$
\begin{equation*}
    = \begin{bmatrix}
        \int_{\dom(\vZ)} \vell(\vy,\vz)_1 \mathrm{d}\vz & \cdots & \int_{\dom(\vZ)} \vell(\vy,\vz)_K \mathrm{d}\vz 
    \end{bmatrix}^\top \!\!\in \bbC^K.
\end{equation*}
Therefore, due to the linearity of the function computed by sum layers, we write
$\int_{\dom(\vZ)} \vW\vell(\vy,\vz)\mathrm{d}\vz =$
\begin{align*}
    &= \vW \begin{bmatrix}
        \int_{\dom(\vZ)} \vell(\vy,\vz)_1 \mathrm{d}\vz & \cdots & \int_{\dom(\vZ)} \vell(\vy,\vz)_K \mathrm{d}\vz
    \end{bmatrix}^\top\\
    &= \vW \int_{\dom(\vZ)} \vell(\vy,\vz) \mathrm{d}\vz.
\end{align*}
For Hadamard product layers in decomposable circuits (\cref{defn:smoothness-decomposability}), we generally write
$\int_{\dom(\vZ_1)\times \dom(\vZ_2)} \vell_1(\vy_1,\vz_1) \odot \vell_2(\vy_2,\vz_2) \mathrm{d}\vz_1\mathrm{d}\vz_2 =$
\begin{align*}
    \int_{\dom(\vZ_1)} \vell_1(\vy_1,\vz_1) \mathrm{d}\vz_1 \odot \int_{\dom(\vZ_2)} \vell_2(\vy_1,\vz_2) \mathrm{d}\vz_2,
\end{align*}
where $(\vY_1,\vY_2)$ is a partitioning of $\vY$ and $(\vZ_1,\vZ_2)$ is a partitioning of $\vZ$.
Similarly, we will write $\int_{\dom(\vZ_1)\times \dom(\vZ_2)} \vell_1(\vy_1,\vz_1) \otimes \vell_2(\vy_2,\vz_2) \mathrm{d}\vz_1\mathrm{d}\vz_2$ as above by replacing $\odot$ with $\otimes$ instead.

\subsection{Normalized Squared Circuits}\label{app:normalized-squared}

\begin{reprop}{prop:orthonormal-circuits-normalization}
    Let $c$ be a structured-decomposable tensorized circuit over variables $\vX$.
    If $c$ is orthonormal, then its squaring encodes a normalized distribution, i.e., $Z=1$.
\begin{proof}
    We prove it by showing how the orthonormal property satisfied by $c$ (\cref{defn:orthonormal-circuit}) yields $Z = 1$.
    In particular, we do so by following \cref{alg:tensorized-square} to compute the modulus square of a tensorized circuit $c$ that is structured-decomposable as yet another smooth and decomposable circuit $c^2$.
    The idea is to recursively show that the output of each layer in $c^2$ must output the flattening (or vectorization) of an identity matrix when computing $Z$, thus yielding $Z=1$ as output in the last step of the recursion.

    \paragraph{Case (i): input layer.}
    Given an input layer $\vell$ computing a vector of $K$ orthonormal functions $\vell(X) = [f_1(X),\ldots,f_K(X)]^\top$, \cref{alg:tensorized-square} materializes another input layer $\vell^2$ such that $\vell^2(X) = \vell(X)\otimes \vell(X)^*$ (L3-4).
    Thus, $\vell^2$ computes a vector of $K^2$ functions $\{f_i(X)f_j^*(X)\mid i,j\in [K]\}$, and it is an input layer of the squared PC $c^2$.
    Since $\vell$ encodes orthonormal functions, we observe that integrating $\vell^2$ over the whole domain of $X$ yields $\int_{\dom(\vZ)} \vell^2(\vz) \mathrm{d}\vz = \vecflat(\vI_K)$, where $\vecflat(\;\cdot\;)$ denotes the flattening of a matrix into a vector.

    \paragraph{Case (ii): Hadamard product layer.}
    Given a Hadamard product layer $\vell$ with scope $\vZ$ and computing $\vell_1(\vZ_1)\odot \vell_2(\vZ_2)$ with $\vZ_1\cap\vZ_2 = \emptyset$, $\vZ_1\cup\vZ_2 = \vZ$, \cref{alg:tensorized-square} constructs another Hadamard product layer $\vell^2$ computing $\vell_1^2(\vZ_1)\odot \vell_2^2(\vZ_2)$, where $\vell_1^2$ (resp. $\vell_2^2$) is the squaring of the layer $\vell_1$ (resp. $\vell_2$) obtained recursively in \cref{alg:tensorized-square} (L8-9).
    Now, if $\int_{\dom(\vZ_1)} \vell_1^2(\vz_1) \mathrm{d}\vz_1 = \vecflat(\vI_{K_{\vell}})$ and $\int_{\dom(\vZ_2)} \vell_2^2(\vz_2) \mathrm{d}\vz_2 = \vecflat(\vI_{K_{\vell}})$, then we have that
    \begin{equation*}
        \int_{\dom(\vZ)} \vell^2(\vz) \mathrm{d}\vz = \vecflat(\vI_{K_{\vell}}) \odot \vecflat(\vI_{K_{\vell}}) = \vecflat(\vI_{K_{\vell}}),
    \end{equation*}
    by exploiting the decomposability of $c$ (thus also $c^2$ \citep{vergari2021compositional}) (\cref{defn:smoothness-decomposability}).

    \paragraph{Case (iii): Kronecker product layer.}
    Given a Kronecker product layer $\vell$ with scope $\vZ$ and computing $\vell_1(\vZ_1)\otimes \vell_2(\vZ_2)$ with $\vZ_1\cap\vZ_2 = \emptyset$, $\vZ_1\cup\vZ_2 = \vZ$, \cref{alg:tensorized-square} constructs a composition of a sum and a Kronecker product layer $\vell^2(\vZ) = \vP (\vell_1^2(\vZ_1) \otimes \vell_2^2(\vZ_2))$, where $\vP$ is a permutation matrix ensuring $\vell^2$ outputs $\vell(\vZ)\otimes \vell(\vZ)$ (as the Kronecker product is not commutative).
    Similarly to the Hadamard product layer above, if we assume by inductive hypothesis that $\int_{\dom(\vZ_1)} \vell_1^2(\vz_1) \mathrm{d}\vz_1 = \vecflat(\vI_{K_{\vell_1}})$ and $\int_{\dom(\vZ_2)} \vell_2^2(\vz_2) \mathrm{d}\vz_2 = \vecflat(\vI_{K_{\vell_2}})$, then we recover that
    \begin{align*}
        \int_{\dom(\vZ)} \vell^2(\vz) \mathrm{d}\vz &= \vP (\vecflat(\vI_{K_{\vell_1}}) \otimes \vecflat(\vI_{K_{\vell_2}})) \\
        &= \vecflat(\vI_{K_{\vell}}),
    \end{align*}
    where $K_{\vell} = K_{\vell_1}\cdot K_{\vell_2}$,
    by again exploiting the decomposability of $c$.

    \paragraph{Case (iv): sum layer.}
    Finally, let $\vell$ be a sum layer over $\vZ$ and computing the matrix-vector product $\vW\vell_1(\vZ)$, with $\vW\in\bbC^{K_1\times K_2}$, $K_1\leq K_2$ and $\vW\vW^\dagger = \vI_{K_1}$ by hypothesis.
    \cref{alg:tensorized-square} materializes a sum layer $\vell^2$ computing $\vell^2(\vZ) = (\vW\otimes \vW^*) \vell_1^2(\vZ)$, where $\vell_1^2$ is the squared layer obtained from $\vell_1$ by recursion (L15).
    Now, if we assume that $\int_{\dom(\vZ)} \vell_1^2(\vz) \mathrm{d}\vz = \vecflat(\vI_{K_2})$, then we have that
    \begin{align*}
        \int_{\dom(\vZ)} \vell^2(\vz) \mathrm{d}\vz &= (\vW\otimes\vW^*) \vecflat(\vI_{K_2}) \\
        &= \vecflat(\vW\vI_{K_2}\vW^\dagger) = \vecflat(\vI_{K_1}).
    \end{align*}
    Therefore, if $\vell$ (resp. $\vell^2$) is the output layer of the tensorized circuit $c$ (resp. $c^2$), then $\vZ=\vX$, $K_1 = 1$ as $\vell$ must output a scalar, and therefore $Z = \int_{\dom(\vX)} |c(\vx)|^2 \mathrm{d}\vx = 1$.
\end{proof}
\end{reprop}

\subsection{A Tighter Marginalization Complexity}\label{app:marginalization}

\begin{algorithm}[t!]
\small
\caption{$\textsc{Marginalize}(c,\vy,\vZ)$}\label{alg:marginalization}
    \textbf{Input:} A structured-decomposable tensorized circuit $c$ over variables $\vX$ that is orthonormal (\cref{defn:orthonormal-circuit}); a set of variables $\vZ$ to marginalize, and an assignment $\vy$ to variables $\vY = \vX\setminus\vZ$. \\
    \textbf{Output:} The marginal likelihood $p(\vy) = \int_{\dom(\vZ)} |c(\vy,\vz)|^2\mathrm{d}\vz$.
\begin{algorithmic}[1]
    \State $\mathsf{out}\leftarrow\mathsf{Map}$
    \Comment{A map from layers $\vell$ in $c$ to their output vector.}
    \State $\mathsf{mar}\leftarrow\mathsf{Map}$
    \Comment{A map from layers $\vell$ to the integral $\int_{\dom(\vZ')} \vell(\vy',\vz') \otimes \vell(\vy',\vz')^* \mathrm{d}\vz'$ with $\vZ' = \uscope(\vell)\cap\vZ$, $\vY'=\uscope(\vell)\cap\vY$}
    \For{$\vell\in\textsc{TopologicalOrdering}(c)$}
        \If{$\uscope(\vell)\subseteq\vZ$} \textsc{Skip} \Comment{Skip to the next layer} \EndIf
        \If{$\vell$ is an input layer}
             \State $\mathsf{out}[\vell]\leftarrow \vell(\vy')$
             \Comment{$\vy'$ denotes $\vy$ restricted to $\vY'$}
             \State \textsc{Skip}
        \EndIf
        \If{$\vell$ is a sum layer with input $\vell_1$ parameterized by $\vW$}
             \If{$\uscope(\vell)\cap\vZ = \emptyset$}
                \Comment{Evaluate $\vell$ \emph{without} squaring}
                \State $\mathsf{out}[\vell] \leftarrow \vW \mathsf{out}[\vell_1]$
            \Else
                \Comment{Evaluate the corresponding squared layer $\vell^2$}
                \State $\mathsf{mar}[\vell] \leftarrow (\vW \otimes \vW^*) \mathsf{mar}[\vell_1]$
            \EndIf
            \State \textsc{Skip}
        \EndIf
        \State
        \Comment{$\vell$ is either an Hadamard or Kronecker product layer}
        \State $(\vell_1,\vell_2)\leftarrow\textsc{GetInputs}(\vell)$
        \If{$\uscope(\vell) \cap \vZ = \emptyset$}
            \Comment{$\vell$ depends on $\vY$ only}
            \State
            \Comment{Evaluate the product layer $\vell$ \emph{without} squaring it}
            \If{$\vell$ is an Hadamard product layer}
                \State $\mathsf{out}[\vell] \leftarrow \mathsf{out}[\vell_1]\odot \mathsf{out}[\vell_2]$
            \Else
                \Comment{$\vell$ is a Kronecker product layer}
                \State $\mathsf{out}[\vell] \leftarrow \mathsf{out}[\vell_1]\otimes \mathsf{out}[\vell_2]$
            \EndIf
            \State \textsc{Skip}
        \EndIf
        \State
        \Comment{$\vell$ is a product layer depending on both $\vY$ and $\vZ$}
        \If{$\uscope(\vell_1)\subseteq\vZ$}
            \Comment{$\vell_1$ depends on $\vZ$ only}
            \State $\vo_1\leftarrow \vecflat(\vI_{K_{\vell_1}})$
            \Comment{$K_{\vell_1}$ denotes the width of $\vell_1$}
        \ElsIf{$\uscope(\vell_1)\cap\vZ = \emptyset$}
            \Comment{$\vell_1$ depends on $\vY$ only}
            \State $\vo_1\leftarrow \mathsf{out}[\vell_1] \otimes \mathsf{out}[\vell_1]^*$
        \Else
            \Comment{$\uscope(\vell_1)$ depends on both $\vY$ and $\vZ$}
            \State $\vo_1\leftarrow \mathsf{mar}[\vell_1]$
        \EndIf
        \State \textbf{repeat} L19-L24 by replacing $\vell_1$ with $\vell_2$ to obtain $\vo_2$
        \State
        \Comment{Evaluate the corresponding squared layer $\vell^2$}
        \If{$\vell$ is an Hadamard product layer}
            \State $\mathsf{mar}[\vell] \leftarrow \vo_1\odot \vo_2$
        \Else
            \Comment{$\vell$ is a Kronecker product layer}
            \State $\mathsf{mar}[\vell] \leftarrow \vP(\vo_1 \otimes \vo_2)$,
            \State \quad where $\vP$ is a permutation matrix (\cref{app:squaring-circuits})
        \EndIf
    \EndFor
    \State \Return $\mathsf{mar}[\textsc{OutputLayer}(c)]$
\end{algorithmic}
\end{algorithm}

\begin{rethm}{thm:marginalization-complexity}
    Let $c$ be a structured-decomposable orthonormal circuit over variables $\vX$.
    Let $\vZ\subseteq \vX$, $\vY=\vX\setminus\vZ$.
    Computing the marginal likelihood $p(\vy) = \int_{\dom(\vZ)} |c(\vy,\vz)|^2 \mathrm{d}\vz$ requires time $\calO(|\phi_{\vY}| S + |\phi_{\vY,\vZ}| S^2)$, where $\phi_{\vY}$ (resp. $\phi_{\vY,\vZ}$) denotes the set of layers in $c$ whose scope depends on only variables in $\vY$ (resp. on variables both in $\vY$ \emph{and} in $\vZ$).
\begin{proof}
    We prove it by constructing \cref{alg:marginalization}, i.e., the algorithm computing the marginal likelihood given by hypothesis.
    \cref{alg:marginalization} is based on two ideas.
    First, integrating sub-circuits whose layer depend only on the variables being integrated over (i.e., $\vZ$) will yield identity matrices, so there is no need to evaluate them.
    Second, the sub-circuits whose layers depend on the variables that are \emph{not} integrated over (i.e., $\vY$) do not need to be squared and can be evaluated without squaring their size (see \cref{app:squaring-circuits}).

    Below, we consider different cases for each layer and based on the variables they depend on.

    \paragraph{Case (i) layers depending on variables $\vZ$ only.}
    Consider a layer $\vell$ in $c$ such that $\uscope(\vell)\subseteq\vZ$, i.e., $\vell\in\phi_{\vZ}$ by hypothesis.
    Moreover, let $\vZ' = \vZ\cap\uscope(\vell)$.
    Since the sub-circuit rooted in $\vell$ is orthonormal by hypothesis, we have that integrating such a sub-circuit yields the flattening of an identity matrix, i.e., $\int_{\dom(\vZ')} \vell^2(\vz') \mathrm{d}\vz' = \vecflat(\vI_K)$, where $\vell^2$ is the squared layer constructed by \cref{alg:tensorized-square}, and $K$ is the size of the outputs of $\vell$.
    This can be seen from our proof of \cref{prop:orthonormal-circuits-normalization}.
    Therefore, layers in $\phi_{\vZ}$ do not need be evaluated, and this is reflected in L4 of \cref{alg:marginalization}.

    \paragraph{Case (ii) layers depending on variables both in $\vY$ and $\vZ$.}
    Consider a layer $\vell$ in $c$ such that $\uscope(\vell)\cap\vY\neq\emptyset$ and $\uscope(\vell)\cap\vZ\neq\emptyset$, i.e., $\vell\in\phi_{\vY,\vZ}$ by hypothesis.
    Moreover, let $\vX' = \uscope(\vell)\subseteq\vX$, $\vZ' = \vX'\cap\vZ$, and $\vY' = \vX'\setminus\vZ'$.
    Since input layers can only be univariate, $\vell$ must be either a sum or product layer.

    Assume $\vell$ is a sum layer in $c$ receiving input from $\vell_1$ and is parameterized by $\vW\in\bbC^{K_1\times K_2}$.
    Then, the corresponding squared layer $\vell^2$ in $c^2$ receives input from $\vell_1^2$ and is parameterized by $\vW\otimes\vW^*$.
    Therefore, we have that
    \begin{equation*}
        \int_{\dom(\vZ')} \vell^2(\vy',\vz') \mathrm{d}\vz' = (\vW\otimes\vW^*) \int_{\dom(\vZ')} \vell_1^2(\vy',\vz') \mathrm{d}\vz',
    \end{equation*}
    hence the integral is simply ``\emph{pushed}'' towards the sub-circuit of $c^2$ rooted in $\vell_1^2$.
    The computation of the above integral for a sum layer can be found at L12 of \cref{alg:marginalization}.

    Now, assume $\vell$ is an Hadamard product layer in $c$ receiving input from $\vell_1,\vell_2$ having scopes $\vX_1',\vX_2'$, respectively.
    Then, the corresponding squared layer $\vell^2$ in $c^2$ is an Hadamard layer receiving inputs from $\vell_1$ and $\vell_2$ (see \cref{app:squaring-circuits}).
    Moreover, let $\vY_1' = \vX_1'\cap \vY$, $\vZ_1' = \vX_1'\cap \vZ$, $\vY_2' = \vX_2'\cap \vY$, $\vZ_2' = \vX_2'\cap \vY$.
    Below we proceed by cases in order to prove L23-36 in \cref{alg:marginalization}.
    Due to decomposability of $c$ (\cref{defn:smoothness-decomposability}), if $\vZ_1'=\emptyset$ and $\vY_2'=\emptyset$ we recover that
    \begin{align*}
        \int_{\dom(\vZ_2')} & \vell^2(\vy_1',\vz_2') \mathrm{d}\vz' = \int_{\dom(\vZ_2')} \vell^2(\vy_1') \odot \vell^2(\vz_2') \mathrm{d}\vz_2' \\
        &= \vell_1^2(\vy_1') \odot \int_{\dom(\vZ_2')} \vell_2^2(\vz_2') \mathrm{d}\vz_2' \\
        &= ( \vell_1(\vy_1') \otimes \vell_1(\vy_1')^* ) \odot \vecflat(\vI_K),
    \end{align*}
    because $\vell_1^2(\vy') = \vell_1(\vy') \odot \vell_1(\vy')^*$ and $\vell_2$ depends on $\vZ$ only (see \textbf{Case (i)}) above.
    Therefore, for this case we do not need to square the sub-circuit rooted in $\vell_1$, i.e., we can evaluate $\vell_1$ as is on the input $\vy_1'$ and then computes the Kronecker product of the output vector with its conjugate only.
    Conversely, if $\vY_1'=\emptyset$ and $\vZ_2'=\emptyset$ we recover a similar result: we do not need to square the sub-circuits rooted in $\vell_2$.
    This case is captured by L24-27 in \cref{alg:marginalization}.

    Furthermore, consider the case $\vZ_1' = \emptyset$, then similarly to the above we recover that
    \begin{align*}
        &\int_{\dom(\vZ_2')} \vell^2(\vy_1',\vz_2') \mathrm{d}\vz' \\
        &= ( \vell_1(\vy_1') \otimes \vell_1(\vy_1')^* ) \odot \int_{\dom(\vZ_2')} \vell_2^2(\vy_2',\vz_2') \mathrm{d}\vz_2'.
    \end{align*}
    Therefore, while we do not require squaring the sub-circuit rooted in $\vell_1$, we however need to square the one rooted in $\vell_2$ and integrate it.
    Conversely, we recover a similar result if $\vZ_2'=\emptyset$.
    This case is captured by L28-29 in \cref{alg:marginalization}.

    If neither $\vZ_1'$ nor $\vZ_2'$ are empty, then we need to square the sub-circuits rooted both in $\vell_1$ and $\vell_2$, since we have that
    \begin{align*}
        &\int_{\dom(\vZ_1'\cup\vZ_2')} \vell^2(\vy_1'\vy,\vz_2') \mathrm{d}\vz' \\
        &= \int_{\dom(\vZ_1')} \vell_1^2(\vy_1',\vz_1') \mathrm{d}\vz_1' \odot \int_{\dom(\vZ_2')} \vell_2^2(\vy_1',\vz_2') \mathrm{d}\vz_2'.
    \end{align*}

    Instead of Hadamard product layers, a similar discussion can be carried for the case of $\vell$ being a Kronecker product layer.
    In particular, the computation of the above integrals for Hadamard or Kronecker product layers for the discussed cases can be found at L23-36 of \cref{alg:marginalization}.

    We now discuss what is the computational complexity for the \textbf{Case (ii)} above.
    First, we observe that we need to quadratically increase the size of each layer that depends on both variables in $\vY$ and in $\vZ$.
    This already requires time $\calO(|\phi_{\vY,\vZ}|S^2)$.
    In addition, we need to compute a Kronecker product of the outputs of layers that depend on variables $\vY$ only, but that are also input to product layers depending on both $\vY$ and $\vZ$.
    However, since (i) each product layer depending on both $\vY$ and $\vZ$ receives input from exactly two other layers $\vell_1,\vell_2$, and (ii) at most one between $\vell_1$ and $\vell_2$ can depend on variables $\vY$ only, we recover that the complexity of computing these Kronecker products is $\calO(|\phi_{\vY,\vZ}|J^2)$, where $J$ is the maximum output size of each layer in $c$.
    Since the output size of each layer $J$ is always bounded by the layer size $S$ (see below \cref{defn:tensorized-circuit}), it turns out that $|\phi_{\vY,\vZ}|J^2 \in \calO(|\phi_{\vY,\vZ}|S^2)$.
    Therefore, the Kronecker products mentioned above account for only a constant multiplicative factor in our complexity.

    \paragraph{Case (iii) layers depending on variables $\vY$ only.}
    In \textbf{Case (ii)} we have shown that we need to evaluate the layers in $c$ whose scope depends on variable $\vY$ only.
    Since such layers do not need to be squared (see above), it turns out computing them requires time $\calO(|\phi_{\vY}|S)$.
    This case is captured by L9-10 and L16-L22 in \cref{alg:marginalization}.

    Therefore, by combining \textbf{Cases (i-iii) above}, we recover the overall time complexity of \cref{alg:marginalization} is $\calO(|\phi_{\vY}|S + |\phi_{\vY,\vZ}|S^2)$.
\end{proof}
\end{rethm}

\subsection{Are Orthonormal Circuits less Expressive?}\label{app:question-expressiveness}

\begin{figure*}[!t]
\begin{subfigure}{0.29\linewidth}
    \centering
\scalebox{0.7}{
\begin{tikzpicture}[cirtikz]
    \node (q) [fill=olive4, anchor=east, minimum width=72pt] {\Prod};
    \node (o) [fill=lacamlilac, anchor=west, minimum width=72pt, right=20pt of q.east] {\Prod};
    \node (r) [fill=olive4, anchor=south, minimum width=55pt, above=20pt of q.north] {\Sum};
    \node (t) [fill=lacamlilac, anchor=south, minimum width=55pt, above=20pt of o.north] {\Sum};
    \node (p) [fill=petroil2, minimum width=55pt, above=20pt of r.north, anchor=south] at ($(r.north)!0.5!(t.north)$) {\Odot};
    \node (s) [fill=petroil2, anchor=south, minimum width=55pt, above=20pt of p.north] {\Sum};
\begin{pgfonlayer}{foreground}
    \node (rw) [anchor=north, below=0pt of r.south] {$\vV_1$};
    \node (tw) [anchor=north, below=0pt of t.south] {$\vV_2$};
    \node (sw) [anchor=north, below=0pt of s.south] {$\vW$};
\end{pgfonlayer}
\begin{pgfonlayer}{background}
    \draw [-, bcedge] (q.north west) -- (r.south west);
    \draw [-, bcedge] (q.north east) -- (r.south east);
    \draw [-, bcedge] (o.north west) -- (t.south west);
    \draw [-, bcedge] (o.north east) -- (t.south east);
    \draw [-, bcedge] (r.north east) -- (t.north west);
    \draw [-, bcedge] (r.north west) -- (p.south west);
    \draw [-, bcedge] (t.north east) -- (p.south east);
    \draw [-, bcedge] (p.north west) -- (s.south west);
    \draw [-, bcedge] (p.north east) -- (s.south east);
\end{pgfonlayer}
\end{tikzpicture}
}
\end{subfigure}%
\hfill
\begin{subfigure}{0.32\linewidth}
    \centering
\scalebox{0.7}{
\begin{tikzpicture}[cirtikz]
    \node (q) [fill=olive4, anchor=east, minimum width=72pt] {\Prod};
    \node (o) [fill=lacamlilac, anchor=west, minimum width=72pt, right=20pt of q.east] {\Prod};
    \node (r) [fill=olive4, anchor=south, minimum width=55pt, above=20pt of q.north] {\Sum};
    \node (t) [fill=lacamlilac, anchor=south, minimum width=55pt, above=20pt of o.north] {\Sum};
    \node (p) [fill=petroil2, minimum width=55pt, above=20pt of r.north, anchor=south] at ($(r.north)!0.5!(t.north)$) {\Odot};
    \node (s) [fill=petroil2, anchor=south, minimum width=55pt, above=20pt of p.north] {\Sum};
\begin{pgfonlayer}{foreground}
    \node (rw) [anchor=north, below=0pt of r.south] {$\vR_1^\dagger\vQ_1^\dagger$};
    \node (tw) [anchor=north, below=0pt of t.south] {$\vR_2^\dagger\vQ_2^\dagger$};
    \node (sw) [anchor=north, below=0pt of s.south] {$\vW$};
\end{pgfonlayer}
\begin{pgfonlayer}{background}
    \draw [-, bcedge] (q.north west) -- (r.south west);
    \draw [-, bcedge] (q.north east) -- (r.south east);
    \draw [-, bcedge] (o.north west) -- (t.south west);
    \draw [-, bcedge] (o.north east) -- (t.south east);
    \draw [-, bcedge] (r.north east) -- (t.north west);
    \draw [-, bcedge] (r.north west) -- (p.south west);
    \draw [-, bcedge] (t.north east) -- (p.south east);
    \draw [-, bcedge] (p.north west) -- (s.south west);
    \draw [-, bcedge] (p.north east) -- (s.south east);
\end{pgfonlayer}
\end{tikzpicture}
}
\end{subfigure}%
\hfill
\begin{subfigure}{0.32\linewidth}
    \centering
\scalebox{0.7}{
\begin{tikzpicture}[cirtikz]
    \node (q) [fill=olive4, anchor=east, minimum width=72pt] {\Prod};
    \node (o) [fill=lacamlilac, anchor=west, minimum width=72pt, right=20pt of q.east] {\Prod};
    \node (r) [fill=olive4, anchor=south, minimum width=55pt, above=20pt of q.north] {\Sum};
    \node (t) [fill=lacamlilac, anchor=south, minimum width=55pt, above=20pt of o.north] {\Sum};
    \node (p) [fill=petroil2, minimum width=72pt, above=20pt of r.north, anchor=south] at ($(r.north)!0.5!(t.north)$) {\Prod};
    \node (s) [fill=petroil2, anchor=south, minimum width=55pt, above=20pt of p.north] {\Sum};
\begin{pgfonlayer}{foreground}
    \node (rw) [anchor=north, below=0pt of r.south] {$\vQ_1^\dagger$};
    \node (tw) [anchor=north, below=0pt of t.south] {$\vQ_2^\dagger$};
    \node (sw) [anchor=north, below=0pt of s.south] {$\vW (\vR_1^\dagger \bullet \vR_2^\dagger)$};
\end{pgfonlayer}
\begin{pgfonlayer}{background}
    \draw [-, bcedge] (q.north west) -- (r.south west);
    \draw [-, bcedge] (q.north east) -- (r.south east);
    \draw [-, bcedge] (o.north west) -- (t.south west);
    \draw [-, bcedge] (o.north east) -- (t.south east);
    \draw [-, bcedge] (r.north east) -- (t.north west);
    \draw [-, bcedge] (r.north west) -- (p.south west);
    \draw [-, bcedge] (t.north east) -- (p.south east);
    \draw [-, bcedge] (p.north west) -- (s.south west);
    \draw [-, bcedge] (p.north east) -- (s.south east);
\end{pgfonlayer}
\end{tikzpicture}
}
\end{subfigure}%
    \caption{\textbf{\cref{alg:orthonormalization} recursively make the sum layer parameter matrices of (semi-)unitary.} Given \emph{a fragment} of a tensorized circuit (left), our algorithm computes QR decompositions of the sum layer parameter matrices $\vV_1^\dagger$ and $\vV_2^\dagger$, thus yielding $\vV_1 = \vR_1^\dagger \vQ_1^\dagger$ and $\vV_2 = \vR_2^\dagger \vQ_2^\dagger$ (mid) (L9-13 in the algorithm).
    The matrices $\vR_1^\dagger,\vR_2^\dagger$ are propagated towards the subsequent Hadamard layer in \cref{alg:orthonormalization}, where $\vR_1^\dagger \bullet \vR_2^\dagger$ is computed (L21) and then multiplied to the parameter matrix $\vW$ (right) (L8).
    Note that the Hadamard product layer is replaced with a Kronecker product layer, accounting for a polynomial increase in the layer size.
    The same procedure is then recursively applied to the parameter matrix $\vW(\vR_1^\dagger \bullet \vR_2^\dagger)$ (not shown).
    }
    \label{fig:orthonormalization-transformation}
\end{figure*}

\begin{rethm}{thm:orthogonalization-algorithm}
    Let $c$ be a tensorized circuit over variables $\vX$.
    Assume that each input layer in $c$ encodes a set of orthonormal functions.
    Then, there exists an algorithm returning an orthonormal circuit $c'$ in polynomial time such that $c'$ is equivalent to $c$ up to a multiplicative constant, i.e., $c'(\vX) = Z^{-\frac{1}{2}} c(\vX)$ where $Z = \int_{\dom(\vX)} |c(\vx)|^2 \mathrm{d}\vx$.
\begin{proof}
    For the proof, we will show the correctness of our \cref{alg:orthonormalization} as to retrieve a tensorized orthonormal circuit $c'$ from $c$ such that $c'(\vX) = \beta c(\vX)$ for a positive real constant $\beta$.
    Before showing this, we observe that, since $c'$ is orthonormal, then
    \begin{equation*}
        p(\vX) = |c'(\vX)|^2 = \beta^2 |c(\vX)|^2.
    \end{equation*}
    Therefore, we must have that $\beta = Z^{-\frac{1}{2}}$ with $Z$ being the partition function of $c^2$, i.e., $Z = \int_{\dom(\vX)} |c(\vx)|^2 \mathrm{d}\vx$.
    In other words, it turns out \cref{alg:orthonormalization} not only returns $c'$, but also the value $\beta = Z^{-\frac{1}{2}}$ thus implicitly computing the partition function.
    It remains to show the correctness of \cref{alg:orthonormalization} as mentioned above.

    Let $c$ be a structured-decomposable tensorized circuit whose input layers encode sets of orthonormal functions.
    We show by structural induction how the orthonormal circuit $c'$ is constructed from $c$ using \cref{alg:orthonormalization}.
    The idea is to apply QR decompositions to the sum layer parameters, retain the (semi-)unitary matrix of the decomposition and ``\emph{push}'' the upper-triangular matrix towards the output layer of the circuit in a bottom-up fashion.
    For this reason, after each recursive step \cref{alg:orthonormalization} also returns a matrix $\vR$, which can be a wide matrix and in general it is not (semi-)unitary.
    In particular, given $(\vell',\vR)$ the output of \cref{alg:orthonormalization} for a layer $\vell$, we associate the semantics $\vell(\uscope(\vell)) = \vR\vell'(\uscope(\vell))$ to it, where the circuit rooted in $\vell'$ is orthonormal by inductive hypothesis.
    Note that for any input layer $\vell$ in $c$ encoding $K$ orthonormal functions, we assume $\vell$ is also in $c'$ and $\vR=\vI_K$ (see L3 of the algorithm).

    \paragraph{Case (i): sum layer.}
    Let $\vell$ be a sum layer with scope $\uscope(\vell) = \vZ$ and computing the matrix-vector product $\vW\vell_1(\vZ)$, with $\vW\in\bbC^{K_1\times K_2}$, $K_1\leq K_2$.
    By applying \cref{alg:orthonormalization} on the circuit rooted in $\vell_1$, we retrieve the layer $\vell_1'$ and the matrix $\vR_1\in\bbC^{K_2\times K_3}$ such that $\vell_1(\vZ) = \vR_1 \vell_1'(\vZ)$ holds.
    Therefore, we can write the function computed by $\vell$ as $\vell(\vZ) = \vW\vR_1 \vell_1'(\vZ)$.
    Let $\vV = \vW\vR_1 \in\bbC^{K_1\times K_3}$, with $K_1\leq K_3$.
    To retrieve a sum layer parameterized by a (semi-)unitary matrix, we apply the QR decomposition on $\vV^\dagger$, i.e.,
    $\vV^\dagger = \vQ\vR$, where $\vQ\in\bbC^{K_3\times K_1}$ and $\vR\in\bbC^{K_1\times K_1}$.
    Here, $\vQ^\dagger\vQ = \vI_{K_1}$ and $\vR$ is an upper triangular matrix.
    Using the QR decomposition above, we can rewrite $\vell(\vZ) = \vR^\dagger\vQ^\dagger \vell_1'(\vZ)$.
    Finally, we retrieve a sum layer $\vell'$ in $c'$ computing $\vell'(\vZ) = \vQ^\dagger \vell_1'(\vZ)$, i.e., $\vell(\vZ) = \vR^\dagger \vell'(\vZ)$.
    Thus, L11 in \cref{alg:orthonormalization} returns both $\vell'$ and $\vR^\dagger$.

    \paragraph{Case (ii): Kronecker product layer.}
    Consider the case $\vell$ is an Kronecker product layer having scope $\vZ = \uscope(\vell) = \uscope(\vell_1) \cup \uscope(\vell_2)$, and computing $\vell(\vZ) = \vell_1(\vZ_1)\otimes \vell_2(\vZ_2)$.
    By inductive hypothesis, let $\vell_1'$ and $\vell_2'$ be the output layers of orthonormal circuits obtained by applying \cref{alg:orthonormalization} on $\vell_1$ and $\vell_2$, respectively.
    Moreover, let $\vR_1\in\bbC^{K_1\times K_2}$ and $\vR_2\in\bbC^{K_3\times K_4}$, with $K_1\leq K_2$ and $K_3\leq K_4$, be the wide matrices returned by the algorithm.
    Therefore, we have that $\vell_1$ (resp. $\vell_2$) computes $\vell_1(\vZ_1) = \vR_1 \vell_1'(\vZ_1)$ (resp. $\vell_1(\vZ_2) = \vR_2 \vell_2'(\vZ_2)$).
    For this reason, we can rewrite the function computed by $\vell$ as
    \begin{align*}
        \vell(\vZ) &= ( \vR_1 \vell_1'(\vZ_1) ) \otimes ( \vR_2 \vell_2'(\vZ_2) ) \\
        &= ( \vR_1 \otimes \vR_2 ) ( \vell_1'(\vZ_1) \otimes \vell_2'(\vZ_2) ),
    \end{align*}
    where we use the Kronecker mixed-product property.
    Finally, we retrieve a Kronecker layer $\vell'$ in $c'$ computing $\vell'(\vZ) = \vell_1'(\vZ_1) \otimes \vell_2'(\vZ_2)$, i.e., $\vell(\vZ) = ( \vR_1 \otimes \vR_2 )\vell'(\vZ)$.
    Thus, L16 in \cref{alg:orthonormalization} returns both $\vell'$ and $\vR_1\otimes \vR_2$.

    \paragraph{Case (iii): Hadamard product layer.}
    Similarly, we consider the case of $\vell$ being an Hadamard product layer having scope $\vZ = \uscope(\vell) = \uscope(\vell_1) \cup \uscope(\vell_2)$, and computing $\vell(\vZ) = \vell_1(\vZ_1)\otimes \vell_2(\vZ_2)$.
    By inductive hypothesis, let $\vell_1'$ and $\vell_2'$ be the output layers of orthonormal circuits obtained by applying \cref{alg:orthonormalization} on $\vell_1$ and $\vell_2$, respectively.
    Moreover, let $\vR_1\in\bbC^{K_1\times K_2}$ and $\vR_2\in\bbC^{K_1\times K_3}$, with $K_1\leq K_2$ and $K_1\leq K_3$, be the wide matrices returned by the algorithm.
    Therefore, we have that $\vell_1$ (resp. $\vell_2$) computes $\vell_1(\vZ_1) = \vR_1 \vell_1'(\vZ_1)$ (resp. $\vell_1(\vZ_2) = \vR_2 \vell_2'(\vZ_2)$).
    For this reason, we can rewrite the function computed by $\vell$ as
    \begin{align*}
        \vell(\vZ) &= ( \vR_1 \vell_1'(\vZ_1) ) \odot ( \vR_2 \vell_2'(\vZ_2) ) \\
        &= ( \vR_1 \bullet \vR_2 ) ( \vell_1'(\vZ_1) \otimes \vell_2'(\vZ_2) ),
    \end{align*}
    where for the last equality we used the Hadamard mixed-product property, and $\bullet$ denotes the face-splitting matrix product operator as defined next.
    \begin{adefn}[Face-splitting product]
        \label{defn:face-splitting-product}
        Let $\vA\in\bbC^{m\times k}$ and $\vB\in\bbC^{m\times r}$ be matrices. The face-splitting product $\vA\bullet\vB$ is defined as the matrix $\vC\in\bbC^{m\times kr}$,
        \begin{equation*}
            \vC = \begin{bmatrix}
                \va_1 \otimes \vb_1 \\
                \vdots \\
                \va_m \otimes \vb_m
            \end{bmatrix} \quad \text{where} \quad \vA = \begin{bmatrix}
                \va_1 \\
                \vdots \\
                \va_m
            \end{bmatrix}
            \quad
            \vB = \begin{bmatrix}
                \vb_1 \\
                \vdots \\
                \vb_m
            \end{bmatrix},
        \end{equation*}
        and $\{\va_i\}_{i=1}^m,\{\vb_i\}_{i=1}^m$ are row vectors.
    \end{adefn}
    Finally, we retrieve a Kronecker layer $\vell'$ in $c'$ computing $\vell'(\vZ) = \vell_1'(\vZ_1) \otimes \vell_2'(\vZ_2)$, i.e., $\vell(\vZ) = ( \vR_1 \bullet \vR_2 )\vell'(\vZ)$.
    Thus, L21 in \cref{alg:orthonormalization} returns both $\vell'$ and $\vR_1\bullet \vR_2$.
    We note that the Hadamard layer is replaced by a Kronecker layer (see e.g. \cref{fig:orthonormalization-transformation}), thus \textbf{Case (iii)} is the only case accounting for a polynomial increase in the size of the layer.

    \textbf{Case (iv): output layer.}
    Finally, we consider the case of $\vell$ being the output layer in $c$, thus resulting in the last step of the recursion.
    W.l.o.g. we consider $\vell$ being a sum layer.
    Then from our \textbf{Case (i)} above, we have that $\vR\in\bbC^{1\times 1}$ is obtained by the QR decomposition of a column vector $\vV^\dagger\in\bbC^{K\times 1}$, thus corresponding to the scalar $r_{11}$ such that $||r_{11} \vV^\dagger||_2 = 1$, i.e., $r_{11} = || \vV^\dagger ||_2^{-1} = \left( \sum_{i=1}^K |v_{i1}|^2 \right)^{-\frac{1}{2}}$.
    Therefore, the non-negative scalar $\beta$ mentioned at the beginning of our proof must be $\beta = r_{11} = Z^{-\frac{1}{2}}$.
    Hence, $Z = \sum_{i=1}^K |v_{i1}|^2$.

    Finally, the computational complexity of \cref{alg:orthonormalization} mainly depends on the complexity of performing QR decompositions and computing Kronecker products of matrices.
    In particular, we need to perform as many QR decompositions as the number of sum layers in $c$, each requiring time $\calO(K_2^3)$ in the case of a matrix $\vV\in\bbC^{K_1\times K_3}$, $K_1\leq K_3$, is $\calO(K_3^3)$.
    In addition, in the worst case $c$ consists of only Kroneckers as product layers, we have to compute Kronecker products between matrices $\vR_1\in\bbC^{K_1\times K_2}$ and $\vR_2\in\bbC^{K_3\times K_4}$, each requiring time $\calO(K_1K_2K_3K_4)$.
    Assuming matrix products require asymptotic cubic time, we recover the overall complexity of \cref{alg:orthonormalization} is $\calO(L_{\mathsf{sum}}J^3 + L_{\mathsf{prod}}J^4)$, where $J$ is the maximum output size of each layer in $c$ and $L_{\mathsf{sum}}$ (resp. $L_{\mathsf{prod}}$) is the number of sum (resp. product) layers in $c$.
\end{proof}
\end{rethm}

\end{document}